\documentclass{article}

\usepackage[final]{neurips_2026}
\usepackage[utf8]{inputenc}
\usepackage[T1]{fontenc}
\usepackage{url}
\usepackage{nicefrac}
\usepackage{amsfonts}
\usepackage{ltablex}   
\usepackage{ragged2e}

\usepackage{microtype}
\usepackage{graphicx}
\usepackage{float}
\usepackage{subcaption}
\usepackage{booktabs} 
\usepackage{longtable}
\usepackage{tabularx}
\usepackage{array}
\usepackage{makecell}
\usepackage{hyperref}
\usepackage{caption} 

\usepackage[table]{xcolor}

\definecolor{rowgray}{RGB}{242,242,242} 
\definecolor{ourred}{RGB}{199,106,122} 
\definecolor{ourblue}{RGB}{240,255,255}

\keepXColumns 

\newcolumntype{Y}[1]{>{\RaggedRight\arraybackslash}p{#1}}
\newcolumntype{C}[1]{>{\centering\arraybackslash}p{#1}}

\usepackage{amsmath}
\usepackage{amssymb}
\usepackage{mathtools}
\usepackage{amsthm}
\usepackage{multirow}
\usepackage{pdflscape}
\usepackage{booktabs}

\usepackage[capitalize,noabbrev]{cleveref}

\theoremstyle{plain}

\theoremstyle{definition}

\theoremstyle{remark}

\usepackage[textsize=tiny]{todonotes}
\newcommand{\xs}[1]{\textcolor{black}{#1}}

\title{OmniEEG-Bench: A Standardized Evaluation Benchmark for EEG Foundation Models}

\author{
\makebox[\textwidth][c]{%
\begin{tabular}{c}
\textbf{Ziling Lu}$^{1}$\textsuperscript{$\dagger$}\quad
\textbf{Zongsheng Li}$^{2,1}$\textsuperscript{$\dagger$}\quad
\textbf{Xinke Shen}$^{1}$\textsuperscript{$\dagger$}\quad
\textbf{Kexin Lou}$^{1,3}$\textsuperscript{$\dagger$}\quad
\textbf{Yingyue Xin}$^{1}$\\
\textbf{Xiaoqi Chen}$^{1}$\quad
\textbf{Shinan Wang}$^{1}$\quad
\textbf{Xiang Chen}$^{1}$\quad
\textbf{Jiahao Fan}$^{1}$\quad
\textbf{Chenyu Huang}$^{1}$\\
\textbf{Xin Xu}$^{1}$\quad
\textbf{Zhoujie Hou}$^{1}$\quad
\textbf{Chen Wei}$^{1,3}$\textsuperscript{*}\quad
\textbf{Quanying Liu}$^{1,3,4}$\textsuperscript{*}\\[3pt]
{\small $^{1}$Department of Biomedical Engineering, Southern University of Science and Technology, Shenzhen, China}\\
{\small $^{2}$School of Computer Science and Engineering, The Chinese University of Hong Kong, Shenzhen, China}\\
{\small $^{3}$Omni-Intelligence, Shenzhen, China}\\
{\small $^{4}$Shenzhen Loop Area Institute, Shenzhen, China}\\[3pt]
{\small \textsuperscript{$\dagger$}Equal contribution}\\
{\small \textsuperscript{*}Corresponding authors: \texttt{liuqy@sustech.edu.cn; chen.wei@omni-intel.cn}}
\end{tabular}%
}
}

\makeatletter
\providecommand{\@trackname}{}
\makeatother

\begin{document}

\maketitle



\begin{figure}[H]
  \centering
  \includegraphics[width=0.8\textwidth]{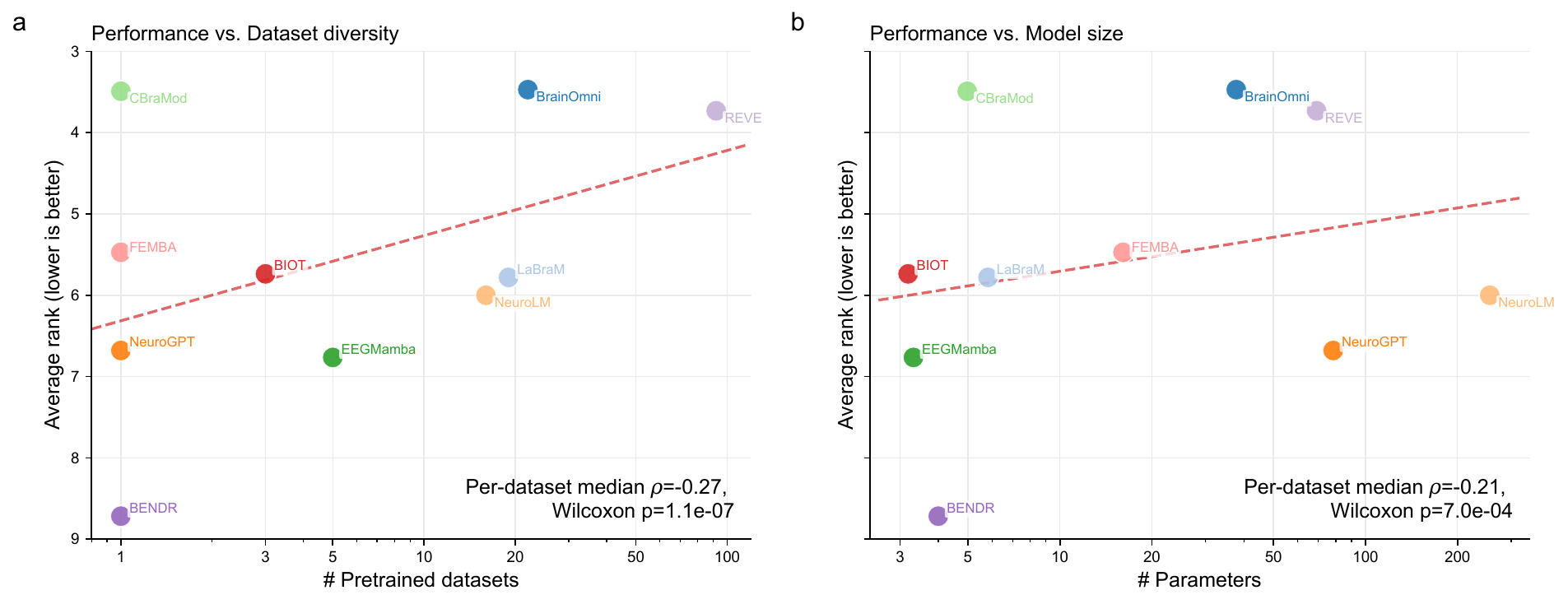}
  \caption{\textbf{Scaling law of pretraining data diversity (a) and model size (b) for linear-probing generalization of EEG foundation models.} Tests on OmniEEG-Bench with 58 datasets show that EEG foundation models pretrained on a greater number of datasets and models with a larger number of parameters tend to achieve lower average ranks (i.e., better performance).}
  \label{fig:scaling_law}
\end{figure}


\begin{abstract}
Electroencephalography (EEG) supports a variety of brain-computer interface (BCI) tasks ranging from brain-state monitoring to human-LLM interactions. EEG foundation models are emerging, but evaluation remains fragmented due to heterogeneous datasets and inconsistent task protocols. Here, we introduce OmniEEG-Bench, a unified benchmark and downstream task roadmap for EEG foundation models (FMs). It organizes evaluation of EEG FMs into six task families spanning (i) signal reliability, (ii) biometrics and disease, (iii) consciousness and state, (iv) cognition and emotion, (v) naturalistic stimulus decoding, and (vi) motor and interaction, introducing a new generation of tasks not systematically benchmarked in prior EEG FM work. OmniEEG-Bench standardizes model deployment, task definitions, and metrics through a task-card specification, and unifies 54 EEG datasets with consistent evaluation protocols. We benchmark 10 representative EEG foundation models and report a leaderboard that covers diverse evaluation settings. Both pretraining dataset diversity and model size are significantly associated with better average ranks across datasets, revealing scaling-law behavior in EEG foundation models (Figure~\ref{fig:scaling_law}). These results suggest that scaling EEG foundation models requires not only larger architectures but also broader and more diverse pretraining data. The benchmark code is available at \url{https://github.com/ncclab-sustech/omni-eegbench.git}.
\end{abstract}


\section{Introduction}

EEG foundation models are rapidly emerging as a new paradigm for brain decoding: by pretraining on large-scale, heterogeneous EEG, a single model can be adapted to many downstream tasks, with the long-term promise of capturing universal EEG representations and enabling practical ``reading the brain''~\cite{kuruppu2025eeg,zhou2025brain,wu2025adabrain}. Recent models, such as BIOT~\cite{yang2023biot}, LaBraM~\cite{jiang2024large} and BrainOmni~\cite{xiao2025brainomni}, explicitly pursue universal and transferable EEG representations via self-supervised pretraining such as masked autoencoding and contrastive learning. However, the field lacks a fair way to compare them: each model is evaluated on different datasets, tasks, and splits in terms of decoding ability and generalizability, and even for the same task, evaluation settings can substantially alter the results. This fragmentation obscures what foundation models truly improve, and hinders the development of robust and generalizable brain decoding systems.

This lack of comparability is not only a protocol issue. It also reflects the absence of a shared view of what EEG foundation models should be good at. Different work implicitly prioritizes different task families, making “general-purpose EEG representation” difficult to operationalize. A coherent task roadmap is therefore a prerequisite for fair evaluation: it makes the capability axes explicit and encourages models to be assessed across a broad spectrum of EEG objectives rather than a few isolated settings. EEG tasks span a wide range of regimes, from clinical abnormality detection~\cite{zhang2023applied} to global state monitoring (such as sleep staging)~\cite{phan2022automatic} and fast, low-latency BCI control~\cite{varbu2022past}. Meanwhile, recent open datasets increasingly capture EEG under more naturalistic, high-dimensional sensory contexts—such as viewing rich visual scenes~\cite{grootswagers2022human} or listening to continuous speech~\cite{broderick2018electrophysiological,chen2025eeg}—where the stimulus space is complex~\cite{sonkusare2019naturalistic}. However, most EEG foundation models are still evaluated on a narrow subset of controlled paradigms, and are rarely tested in a standardized way on these naturalistic-context datasets. Therefore, a benchmark should reflect the field’s evolution from tightly controlled experiments toward naturalistic, interaction-centric paradigms, while maintaining standardization through well-defined tasks.


We introduce OmniEEG-Bench, a standardized evaluation benchmark for EEG foundation models. OmniEEG-Bench organizes downstream evaluation into six task families: (i) signal reliability, (ii) biometrics and disease, (iii) consciousness and state, (iv) cognition and emotion, (v) naturalistic stimulus decoding, and (vi) motor and interaction. This taxonomy is organized by three dimensions: from stable traits to transient states, from slow- to fast-changing temporal scales, and from passive monitoring to active interaction. We standardize preprocessing, task definitions, and metrics through a task-card specification. We benchmark 10 EEG foundation models on 54 EEG datasets with a protocol suite that probes four complementary capabilities: (i) multi-subject cross-trial adaptation, (ii) cross-subject transfer, (iii) label-efficiency via zero-/few-shot adaptation, and (iv) noise robustness under sensor degradation via channel corruption. Our contributions are threefold:

\begin{itemize}
  \item \textbf{A unified task roadmap for EEG foundation models.} We organize EEG downstream evaluation into six task taxonomies and formalize each task with a \emph{task-card} specification that standardizes preprocessing, inputs/outputs, and metrics.
  \item \textbf{Standardized evaluation protocols that probe transfer, data efficiency, and robustness.} We report results under multi-subject (trial-level splits with pooled subjects) and cross-subject (held-out subjects) settings, and further include zero-shot/few-shot adaptation and channel corruption tests. These tests systematically characterize the transferability and robustness of pretrained representations.
  \item \textbf{Large-scale, reproducible benchmarking with diagnostic insights.} We benchmark 10 EEG foundation models on 54 EEG datasets, release a public leaderboard, and \xs{identify the predictive factors that influence the model's downstream performance}.
\end{itemize}

\begin{figure*}[t]
  \centering
  \includegraphics[width=\textwidth]{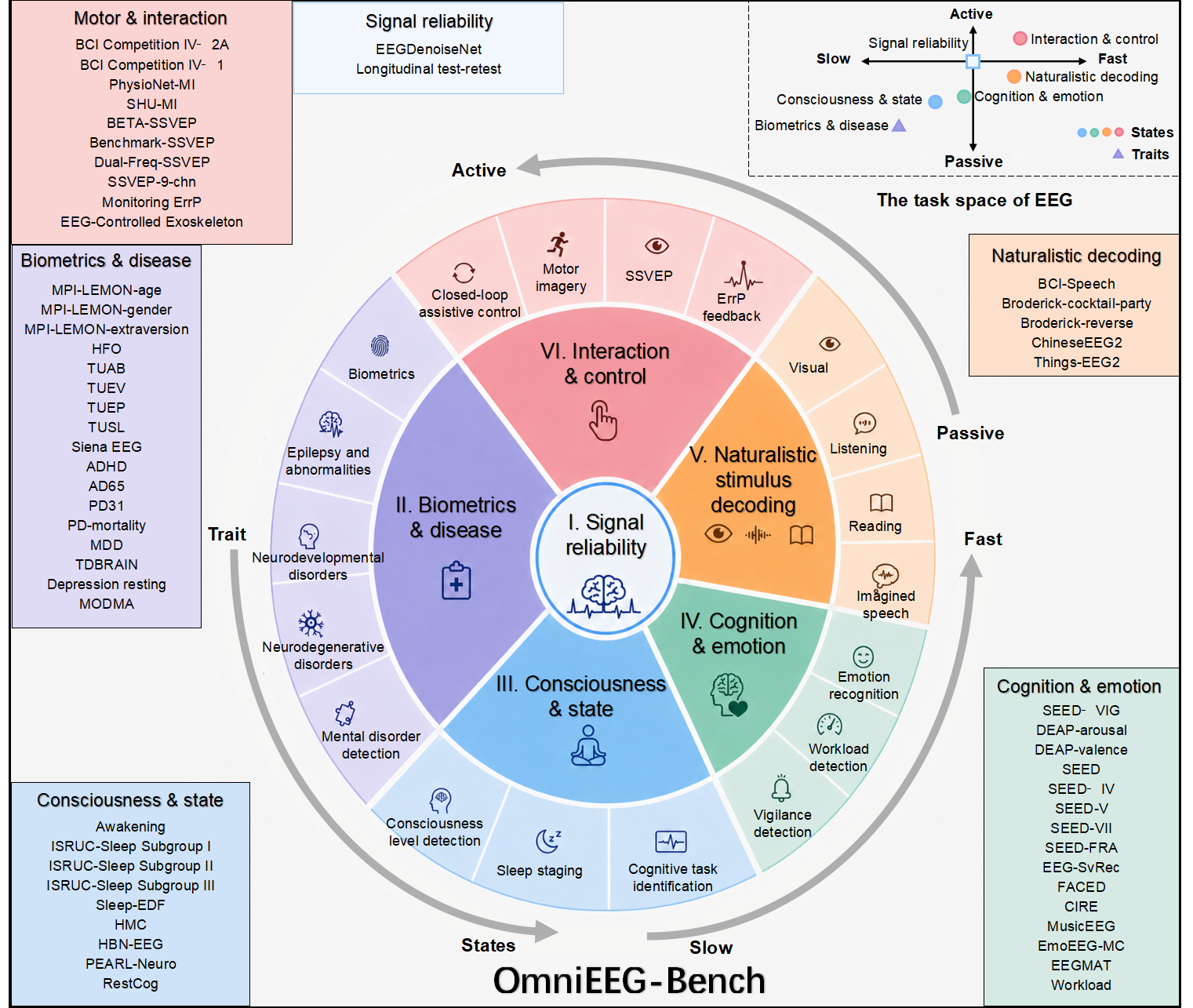}
  \caption{Task taxonomy of OmniEEG-Bench. We organize 58 tasks (from 54 datasets) into 6 categories: signal reliability, biometrics and disease, consciousness and state, cognition and emotion, naturalistic stimulus decoding, and motor and interaction.}
  \label{fig:datasets}
\end{figure*}

\section{EEG Task Taxonomy and Dataset Organization}
\label{sec:taxonomy}

OmniEEG-Bench adopts a capability-driven task taxonomy that organizes downstream EEG tasks into a continuous task space, as illustrated in Fig.~\ref{fig:datasets}. We use three broad organizing dimensions to guide this design. First, tasks differ in the extent to which they reflect stable \textbf{subject-specific characteristics} or \textbf{transient brain states}, consistent with the view that EEG measurements can contain both trait-like and state-dependent components~\cite{martin2025states}. Biometrics and clinical phenotyping emphasize relatively stable individual variability, whereas sleep staging, vigilance, emotion recognition, and task-context decoding emphasize time-varying neural dynamics. Second, tasks differ in \textbf{temporal scale}, ranging from slow global state estimation to faster perceptual, cognitive, and interaction-related decoding. This dimension separates tasks that can be characterized over extended windows from those requiring more temporally precise predictions. Naturalistic stimulus decoding is included as an important dynamic regime because naturalistic stimuli provide richer and more continuous sensory contexts than classical controlled paradigms~\cite{sonkusare2019naturalistic}. Third, tasks differ in their \textbf{interaction regime}, from passive monitoring to active brain-computer interaction. In passive settings, EEG is used to monitor cognitive, affective, or brain-state variables without requiring the user to actively generate control signals. In active settings, the user intentionally modulates mental activity, such as motor imagery, and model predictions may support closed-loop control~\cite{gao2021interface}.

Under this framing, the six OmniEEG-Bench families cover complementary task regions of EEG: \textbf{signal reliability} assesses artifact sensitivity and session consistency, \textbf{biometrics and disease} captures trait-like individual and clinical variability, \textbf{consciousness and state} targets global brain-state dynamics, \textbf{cognition and emotion} covers affective and cognitive changes at intermediate time scales, \textbf{naturalistic stimulus decoding} evaluates perceptual and semantic decoding under dynamic real-world stimuli, and \textbf{motor and interaction} focuses on time-sensitive intent decoding and control.

\paragraph{Type-I: Signal reliability.}
This family probes artifact-aware and session-consistent representations. We include (i) ocular artifact/noise identification (EEGDenoiseNet) and (ii) longitudinal stability, where representations of the same participant should remain consistent across sessions (Longitudinal test-retest).

\paragraph{Type-II: Biometrics and disease.}
This family evaluates clinically and biologically meaningful individual differences. Subtypes include (i) stable traits (MPI-LEMON derived age/gender/personality splits), (ii) epilepsy and abnormalities (HFO, TUAB, TUEP, TUSL, TUEV, and Siena EEG), (iii) neurodevelopmental disorders (ADHD), (iv) neurodegenerative disorders (AD65, PD31, and PD mortality prognosis), and (v) mental disorders (MDD, TDBRAIN, Depression resting, and MODMA).

\paragraph{Type-III: Consciousness and state.}
This family targets global brain state and slow-to-intermediate dynamics. Subtypes include (i) consciousness level detection (Awakening), (ii) sleep staging (e.g., ISRUC-Sleep I/II/III, Sleep-EDF, and HMC), and (iii) cognitive task identification (e.g., HBN-EEG, PEARL-Neuro, RestCog).

\paragraph{Type-IV: Cognition and emotion.}
This family covers affective and cognitive labels with more rapid dynamics. We include (i) vigilance detection (SEED-VIG), (ii) emotion recognition across diverse elicitation settings and label granularities, ranging from video-driven affect induction (DEAP, SEED, SEED-IV, SEED-V, SEED-VII, and SEED-FRA, EEG-SVRec, FACED) to music (MusicEEG) and conversational (CIRE) contexts, with label spaces spanning binary valence or arousal classification (DEAP), mid-granularity categorical emotion (SEED/SEED-IV/SEED-V/SEED-VII, 3--7 classes), and more fine-grained categories (FACED, 9 classes), and (iii) cognitive load assessment (EEGMAT and Workload).

\paragraph{Type-V: Naturalistic stimulus decoding.}
This family targets fast-timescale decoding under naturalistic stimuli, where labels correspond to stimulus attributes or natural listening/reading/viewing context. We include (i) natural speech perception and auditory attention, encompassing multi-class speech phrase discrimination (BCI Speech), selective attention to competing speakers (Broderick, cocktail party), and forward versus reversed speech discrimination (Broderick, reverse); (ii) natural reading with tonal discrimination (ChineseEEG2 reading aloud condition, four Mandarin tones); and (iii) visual semantic categorization (ThingsEEG2, biological vs. non-biological).

\paragraph{Type-VI: Motor and interaction.}
This family focuses on active intent decoding for control and interaction-centric monitoring. Subtypes include (i) motor imagery (BCIC-IV-2a, BCIC-IV-1, PhysioNet-MI, and SHU-MI), (ii) SSVEP control (BETA-SSVEP, Benchmark-SSVEP, Dual-Freq-SSVEP, SSVEP-9-chn), (iii) error-related potential feedback decoding (Monitoring ErrP), and (iv) closed-loop assistive control (EEG-controlled exoskeleton).

See Appendix \ref{sec:task_details} for details and references of the 58 tasks in OmniEEG-Bench.

\begin{figure*}[t]
  \centering
  \includegraphics[width=0.9\textwidth]{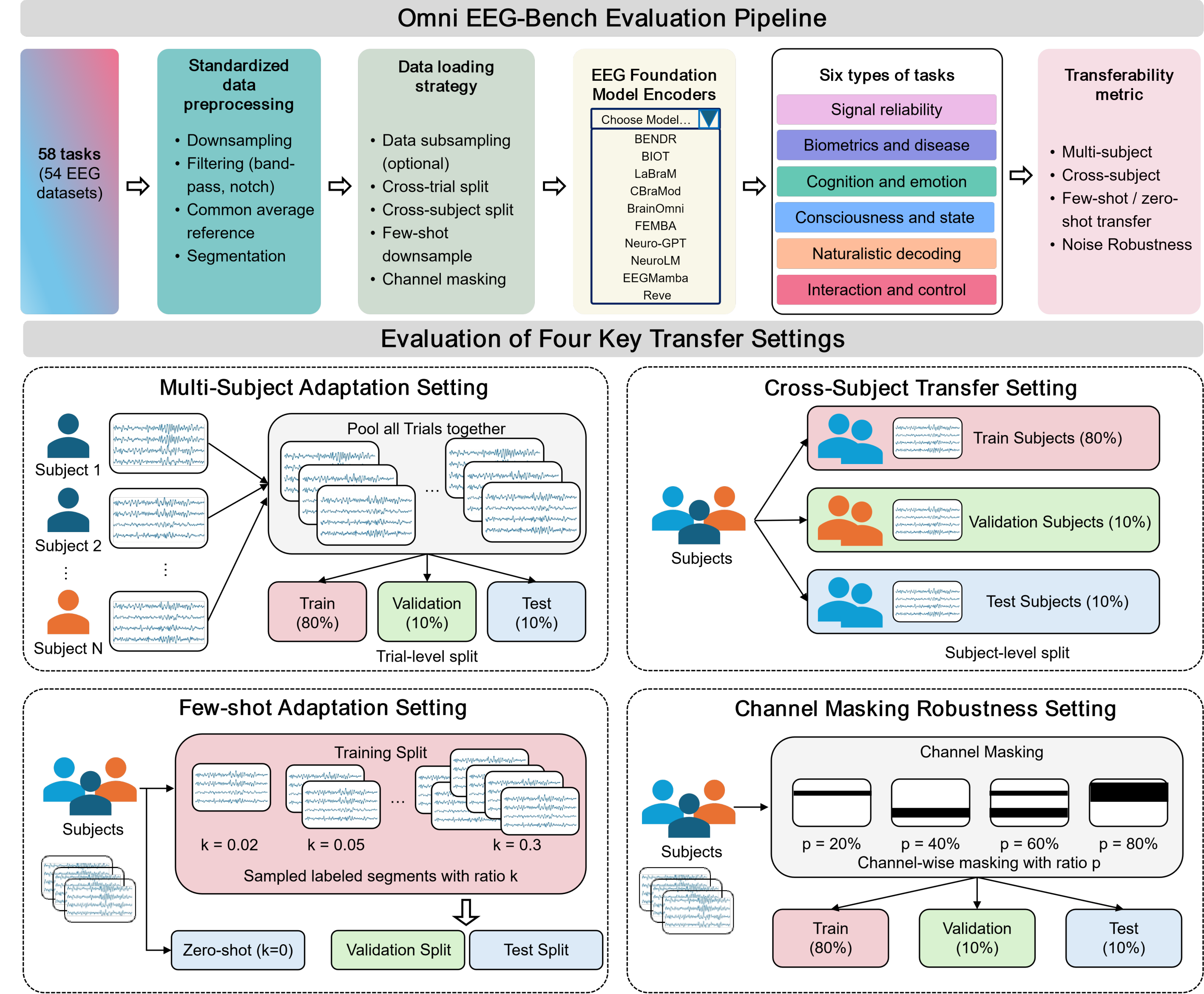}
  \caption{OmniEEG-Bench evaluation pipeline, equipped with four evaluation protocols: cross-subject transfer, multi-subject trial-level adaptation, zero-/few-shot adaptation, and channel-masking robustness.}
  \label{fig:pipeline}
\end{figure*}

\section{Evaluation Protocols}
\label{sec:protocols}

\begin{figure*}[!t]
  \centering
  \includegraphics[width=1.03\textwidth]{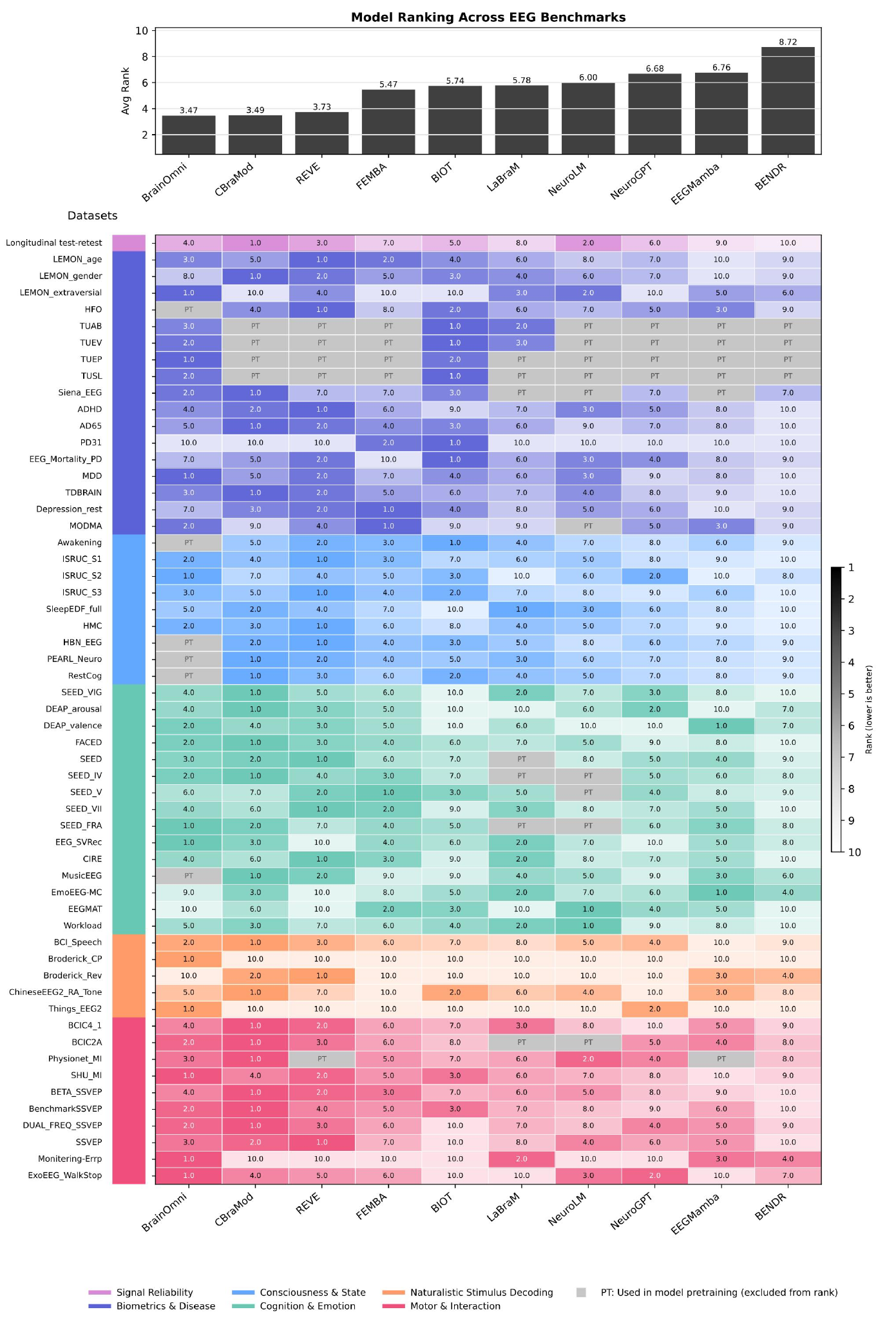}
  \caption{Primary cross-subject-prioritized leaderboard of ten EEG foundation models.}
  \label{fig:cross_subject}
\end{figure*}

To systematically assess the transferability, data efficiency, and robustness of EEG foundation models, we implement four complementary evaluation paradigms: (i) cross-subject transfer, (ii) multi-subject adaptation, (iii) few-shot adaptation, and (iv) channel masking robustness (Fig.~\ref{fig:pipeline}). These configurations are materialized through flexible data loading strategies, including cross-subject splits, cross-trial splits, few-shot downsampling, and channel masking. We preprocess the 54 EEG datasets through a standardized pipeline of downsampling, band-pass and notch filtering, common average referencing, and window segmentation (See Appendix~\ref{sec:preproc} for details). To facilitate efficient benchmarking while maintaining statistical reliability, we randomly select up to 40 samples per subject per class for linear probing (determined by variance stabilization analysis in Appendix~\ref{sec:downsample_validity}), from which all subsequent splits derive.

All backbone architectures are wrapped in a common interface that accepts an EEG sample $\boldsymbol{x} \in \mathbb{R}^{C \times T}$ and outputs a representation $\boldsymbol{z} = f_\theta(\boldsymbol{x})$. A lightweight linear classifier $g(\cdot)$ maps $\boldsymbol{z}$ to task logits. We adopt linear probing as the primary evaluation method---freezing the pretrained backbone $\theta$ and training only the classification head. This protocol enables heterogeneous models to be evaluated under consistent input specifications and optimization settings. To characterize the performance ceiling of each architecture, we additionally perform full fine-tuning on all the datasets from each task category under the cross-subject setting. We benchmark ten representative EEG foundation models: BENDR, BIOT, LaBraM, CBraMod, BrainOmni, FEMBA, Neuro-GPT, NeuroLM, EEGMamba, and REVE. We report results averaged over multiple independent runs, each using a pre-generated, fixed data split derived from a different random seed. Three runs were employed for cross-subject transfer and multi-subject adaptation, and five for zero-/few-shot adaptation and channel masking. This ensures fair and comparable performance estimates in the benchmark. The four evaluation protocols are detailed below:

\paragraph{Cross-subject transfer and primary leaderboard.}
In the \textit{cross-subject} setting, splits are made at the subject level: subjects are partitioned into train/validation/test groups with a ratio of 8:1:1. For the primary leaderboard, we use cross-subject transfer whenever subject-level splitting is meaningful. For the signal-reliability task (Longitudinal test-retest) that do not admit a standard held-out-subject formulation, we use the corresponding multi-subject protocol as a fallback and include them in the main leaderboard for task coverage.

\paragraph{Multi-subject adaptation.}
In the \textit{multi-subject} setting, samples from each subject are split into train/validation/test with an 8:1:1 ratio over trials and pooled across subjects. This setting requires the model to generalize to unseen trials, measuring adaptation when training data spans multiple subjects. 

\paragraph{Zero-shot and few-shot adaptation.}
To quantify data efficiency, we evaluate \textit{few-shot} adaptation by sampling labeled examples of ratio $k$ per class from the training split ($k \in \{0.02, 0.05, 0.1, 0.3\}$). \textit{Zero-shot} is treated as the $k=0$ special case with no dataset-specific supervised training. We compute the pairwise cosine similarity between sample embeddings. For each sample from the test split, we identify its most similar sample from the validation split (to avoid information leakage) in the embedding space and check whether the two samples share the same class label. The prediction is counted as correct if the labels match and incorrect otherwise. The data split follows the cross-subject setting.

\paragraph{Channel masking robustness.}
To measure the model's robustness to missing or degraded sensors, we apply \textit{channel masking} in linear probing:
For each sample, a random subset of channels is zero-masked with a corruption ratio
$p \in \{20\%,40\%,60\%,80\%\}$.
We report the performance with increasing channel corruptions, using fixed corruption seeds across models. The data split also follows the cross-subject setting.

\section{Results}

\subsection{Primary cross-subject transfer benchmark}

In the primary linear-probing benchmark, we prioritize cross-subject transfer because EEG foundation models are expected to learn representations that generalize to unseen individuals. For the dataset where cross-subject evaluation is not meaningful or not applicable (Longitudinal test-retest), we use the corresponding multi-subject evaluation as a fallback. Under this cross-subject-prioritized protocol, BrainOmni achieves the best overall average rank, followed by CBraMod and REVE (Fig.~\ref{fig:cross_subject}; see Supplementary Table 5 for detailed accuracy).


Across tasks, the majority of models achieve above-chance performance. Several tasks are especially difficult, including Parkinson's detection (PD31), arousal classification (DEAP-arousal), speech attention detection (Broderick-Cocktail-party), natural-versus-reversed speech classification (Broderick-reverse), image concept identification (ThingsEEG2), and error-related potential detection (Monitoring-Errp). Overall, naturalistic stimulus decoding emerges as one of the most challenging task categories for current EEG foundation models.

Under full fine-tuning, the model rankings change substantially compared with linear probing, with CBraMod, LaBraM, and FEMBA achieving the top three overall ranks with average ranks of 4.51, 4.88, and 5.42, respectively (Supplementary Fig. 3; see Supplementary Table 7 for detailed accuracy). This suggests that these models can benefit substantially from task-specific end-to-end optimization. We further compare EEG foundation models with two task-specific baselines, EEGConformer and EEGNet. Under full fine-tuning, seven foundation models outperform EEGConformer (avg. rank 7.25) and nine outperform EEGNet (avg. rank 8.24), highlighting the advantage of pretrained models after task-specific adaptation. In contrast, under linear probing, only five foundation models surpass EEGConformer (avg. rank 6.66), while five fall behind this baseline (Supplementary Fig. 5), indicating that frozen pretrained representations remain substantially limited for direct transfer.

\subsection{Zero-shot and few-shot learning}

\begin{figure*}
  \centering
  \includegraphics[width=\textwidth,height=0.3\textheight,keepaspectratio]{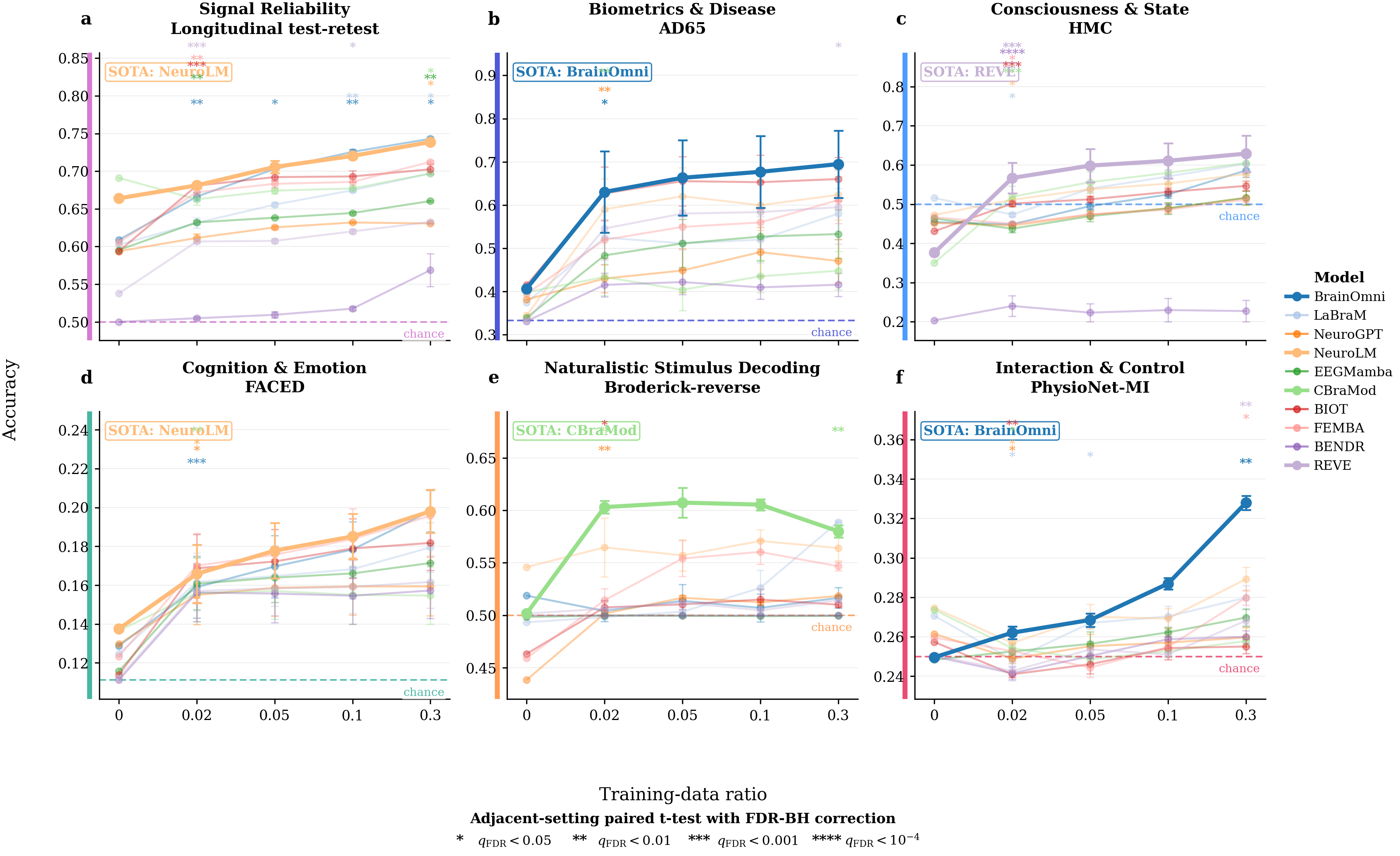}
  \caption{The performance of zero-shot and few-shot learning.}
  \label{fig:few_shot}
\end{figure*}

Under few-shot adaptation, BrainOmni exhibits steeper performance scaling with sample size, indicating superior sample efficiency during linear probing (Fig.~\ref{fig:few_shot}). Task categories display heterogeneous scaling behaviors: Longitidinal test-retest and HMC datasets show gradual, monotonic gains as training data increases. By contrast, most models plateau in FACED, AD65, and Physionet-MI datasets once the few-shot ratio exceeds 0.05. Notably, several models like BrainOmni defy this saturation trend in FACED as well as Physionet-MI datasets, continuing to improve beyond this threshold.

\subsection{Robustness to channel corruptions}

\begin{figure*}
  \centering
  \includegraphics[width=\textwidth,height=0.3\textheight,keepaspectratio]{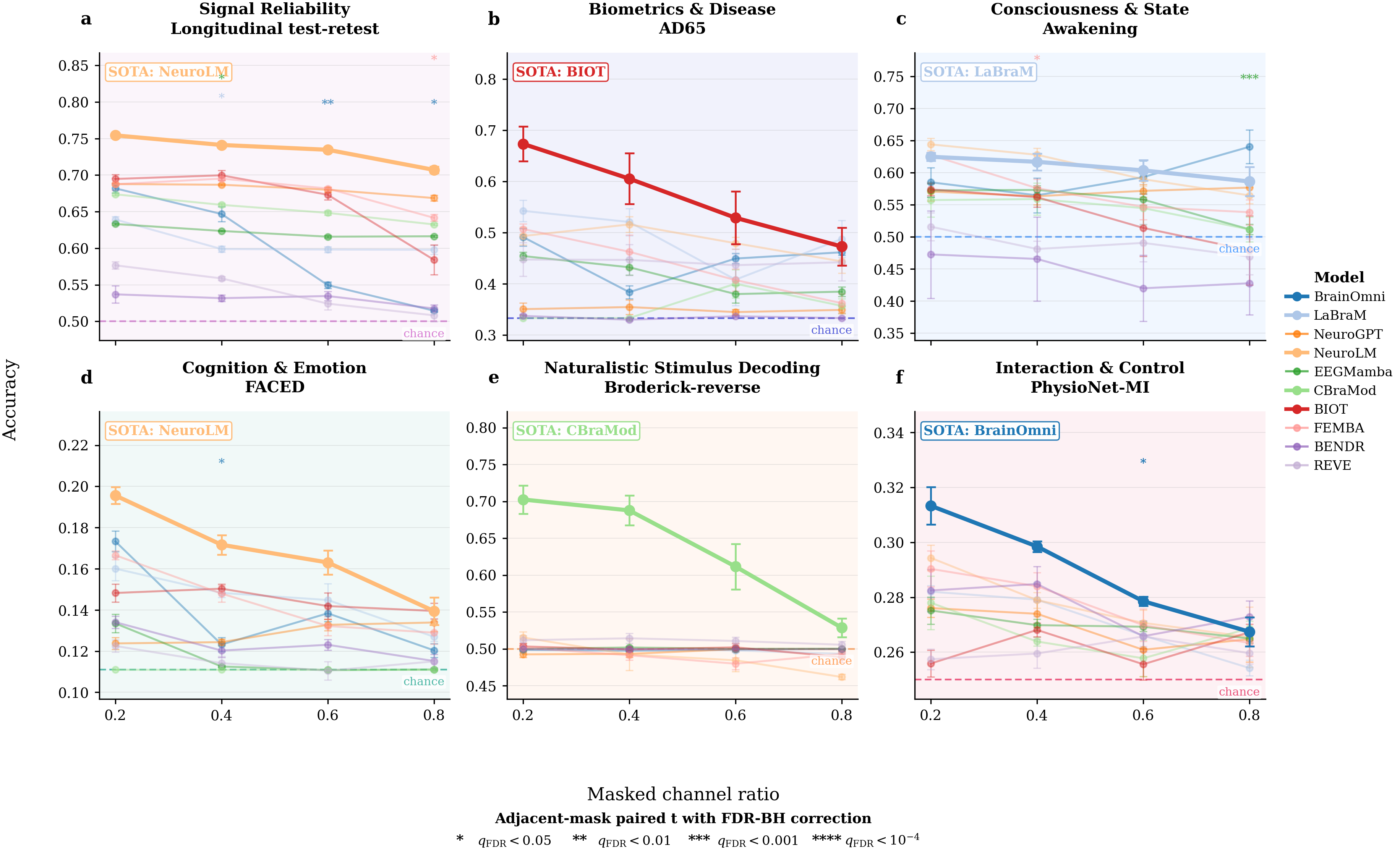}
  \caption{The robustness of model performance on channel masking.}
  \label{fig:channel_mask}
\end{figure*}

On the longitudinal test-retest dataset, most models exhibit robustness to channel corruption, except for BrainOmni (Fig.~\ref{fig:channel_mask}). On other tasks, performance degrades gradually as channel dropout increases. Notably, BIOT sustains stable performance under moderate corruption (ratios of 0.2 and 0.4). At severe corruption levels (ratios of 0.6 and 0.8), most models collapse to near-chance performance.

\subsection{Predictive factors of model performance}

To identify the key factors that influence the generalization performance of EEG foundation models, we systematically analyzed the relationships between model performance and the number of pretraining datasets, number of training subjects, training hours, model size, publication year, model architecture, pretraining paradigm, tokenization strategy, and spatial modeling design. First, Fig.~\ref{fig:scaling_law} shows that both the number of pretraining datasets and model size are significantly associated with better average ranks across datasets. Specifically, models pretrained on more datasets tend to achieve lower average ranks, suggesting that dataset diversity is an important factor for improving generalization. Meanwhile, larger models also show better overall rankings, indicating emerging scaling-law behavior in EEG foundation models. Further per-dataset Spearman correlation analyses support this trend. For each dataset, we computed the Spearman correlation between each model factor and the model rank, and then used a Wilcoxon signed-rank test to assess whether the resulting correlations were systematically different from zero. Because lower ranks indicate better performance, a negative correlation means that a larger value of the corresponding factor is associated with better model performance. The number of pretraining datasets shows a median correlation of $\rho=-0.27$ with a Wilcoxon $p=1.1\times10^{-7}$, indicating that the association between dataset diversity and improved performance is consistent across multiple datasets.

\begin{figure*}
  \centering
  \includegraphics[width=\textwidth]{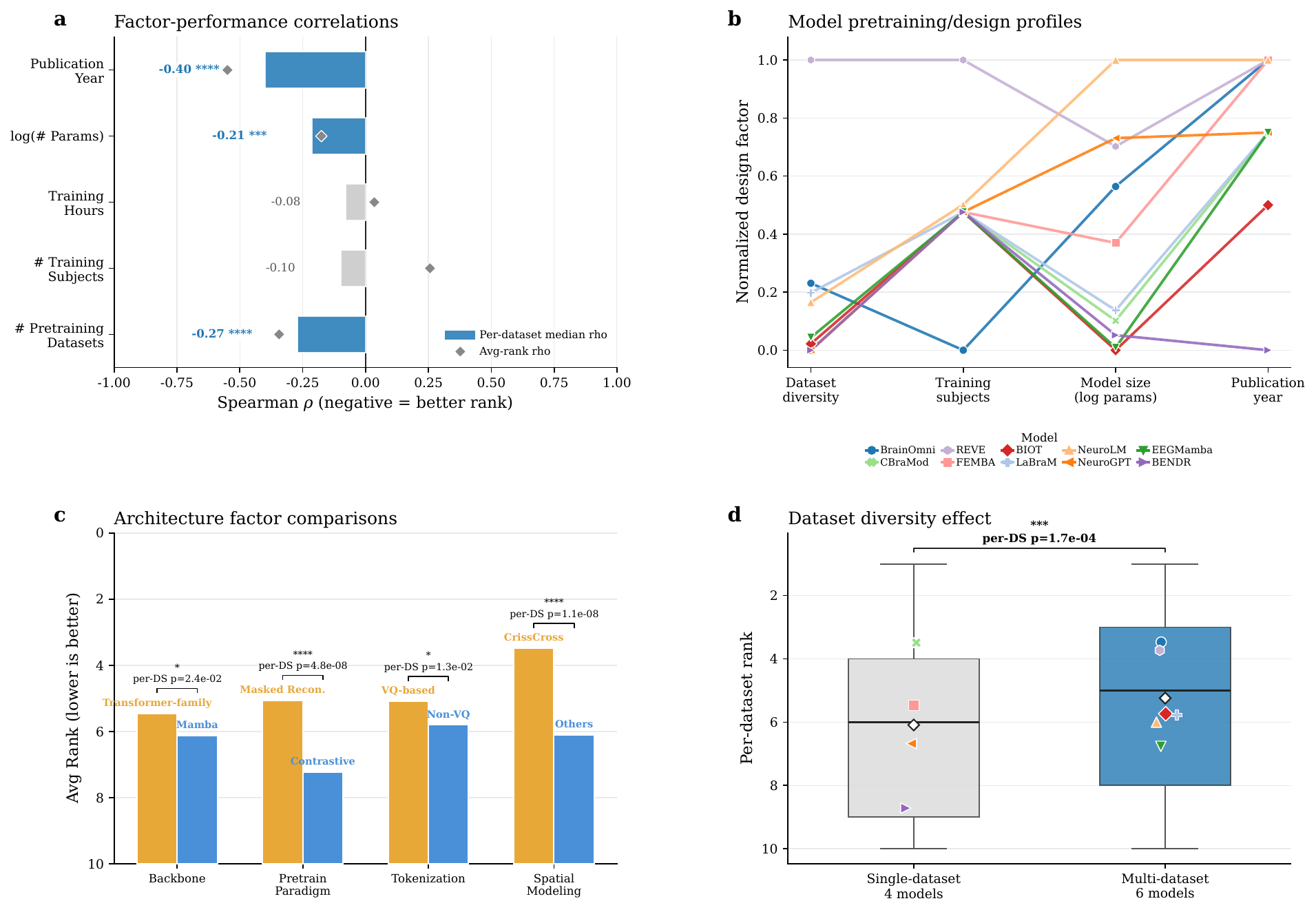}
  \caption{Pretraining design factors and cross-subject linear-probing performance of EEG foundation models. Quantitative factors are evaluated via per-dataset Spearman correlations with Wilcoxon signed-rank tests (a, b). Per-dataset median rho represents the median correlations of model ranks and the predictive factor on each dataset. Avg-rank rho represents the correlation of average model ranks and the predictive factor. Qualitative factors are compared using paired Wilcoxon tests between model groups (c), and single- versus multi-dataset pretraining is compared via Mann--Whitney U test (d).}
  \label{fig:analysis}
\end{figure*}

Fig.~\ref{fig:analysis} further compares the effects of different model factors on performance. Among quantitative factors, publication year, model size, and the number of pretraining datasets are all significantly associated with better per-dataset ranks, with publication year showing the strongest correlation. This suggests that more recent EEG foundation models tend to benefit from larger model capacity, richer pretraining data, and updated modeling designs. In contrast, training hours and the number of training subjects show weaker associations with performance, suggesting that simply increasing training time or the number of subjects does not necessarily lead to stable improvements in downstream generalization. Qualitative architecture analyses show that masked reconstruction pretraining, VQ-based tokenization, and Criss-Cross spatial modeling each significantly outperform their respective alternatives, indicating that effective architectural and training-objective designs remain critical determinants of performance. Finally, models pretrained on multiple datasets outperform those pretrained on a single dataset overall, further supporting the importance of dataset diversity for EEG foundation model generalization. Overall, these results indicate that improvements in EEG foundation models are not driven by a single factor, but by the joint effects of pretraining data diversity, model scale, temporal progress in model development, and architectural design.

\section{Conclusion}

In this work, we present OmniEEG-Bench, a comprehensive benchmark for evaluating EEG foundation models across 54 datasets and 58 tasks. By organizing downstream EEG evaluation into six task families and standardizing preprocessing, task definitions, and evaluation protocols, OmniEEG-Bench provides a unified testbed for assessing the transferability, data efficiency, and robustness of EEG foundation models.

Our results show that EEG foundation models can benefit substantially from task-specific fine-tuning, whereas frozen representations remain limited under direct linear probing. Performance also varies markedly across task families, with naturalistic stimulus decoding posing one of the greatest challenges for current models. Beyond the leaderboard, our diagnostic analyses reveal scaling-law-like trends: models pretrained on more diverse datasets and models with larger parameter counts tend to achieve better average ranks across datasets. These findings suggest that improving EEG foundation models may require not only larger architectures, but also broader and more diverse pretraining data. In addition, VQ-based tokenization and criss-cross spatial modeling are associated with stronger downstream performance, indicating that architectural and pretraining-design choices remain critical for generalization.

By providing a public leaderboard, standardized task cards, and transparent evaluation protocols, OmniEEG-Bench aims to support more reproducible and comparable research on EEG foundation models. We hope that this benchmark will serve as a foundation for tracking progress, diagnosing model limitations, and guiding the development of more generalizable EEG representation learning methods.


\section{Limitations}

In this work, the evaluation is limited to linear probing and full fine-tuning without exploring parameter-efficient transfer strategies. Additionally, the 54 datasets, while diverse, do not exhaustively cover all clinical populations, recording contexts, or geographic regions, and the channel corruption simulation simplifies real-world sensor degradation patterns. 

\section*{Impact Statement}

The goal of this paper is to advance the field of brain-computer interfaces through standardized benchmarking of EEG foundation models. EEG data have diverse applications in clinical monitoring, assistive technologies, and neuroscience research, and our work aims to support progress in these areas through systematic, rigorous evaluations. There are potential societal consequences of improvements in EEG-based modeling, including enhanced diagnostic tools and accessible neurotechnologies. However, these developments also raise considerations around data privacy, equitable access, and responsible deployment. Although our benchmark itself does not introduce new modeling techniques, it can accelerate research that impacts users across clinical and consumer domains. We encourage careful consideration of ethical, privacy, and societal implications in subsequent research that builds upon this benchmark.






\clearpage
\small
\bibliographystyle{unsrt}  
\bibliography{example_paper.bib}

\newpage
\appendix
\onecolumn

\section{Tasks and datasets}
\label{sec:task_details}
This supplementary section provides a complete inventory of the datasets included in OmniEEG-Bench. Supplementary Table 1 summarizes, for each dataset, its assigned taxonomy and subtype, the number of subjects and channels, and a concise description of the corresponding classification objective. See references for detailed information on the datasets.

\captionsetup{labelformat=empty}
\captionsetup{labelformat=empty}
 
\begingroup
\setlength{\tabcolsep}{2.5pt}
\renewcommand{\arraystretch}{1.08}
\scriptsize
\sloppy
\emergencystretch=1em
 
\newcolumntype{P}[1]{>{\RaggedRight\arraybackslash}p{#1}}
\newcolumntype{C}[1]{>{\centering\arraybackslash}p{#1}}
 
\begin{longtable}{@{}
P{1.75cm}
P{3.05cm}
C{0.85cm}
C{0.85cm}
C{1.10cm}
P{\dimexpr\textwidth-1.75cm-3.05cm-0.85cm-0.85cm-1.10cm-10\tabcolsep\relax}
@{}}
\caption{\textit{Supplementary Table 1}. OmniEEG-Bench dataset inventory.}
\label{tab:omnieeg_datasets}\\
 
\toprule
\rowcolor{rowgray}
Task type & Datasets & \#Subjects & \#Channels & Sample length & Task description \\
\midrule
\endfirsthead
 
\toprule
\rowcolor{rowgray} Task type & Tasks & \#Subjects & \#Channels & Sample Length & Task description \\
\midrule
\endhead
 
\bottomrule
\endlastfoot
 
  \rowcolor{ourblue} \multicolumn{6}{l}{\textbf{Type-I: Signal reliability}}\\
  \cline{1-6}
  \multirow{1}{=}{Artifact identification}
  & 1. EEGDenoiseNet~\cite{zhang2021eegdenoisenet}
  & 1 & 1 & 2 s & Single-channel noise-related binary classification (2 classes).\\
  \cline{1-6}
  \multirow{1}{=}{Test-retest reliability}
  & 2. Longitudinal test-retest~\cite{ds006940:1.0.0}
  & 45 & 60 & 2 s & Cross-session subject identification (2 classes in the current mounted version).\\
  \midrule
  \addlinespace[3pt]
 
  \rowcolor{ourblue}\multicolumn{6}{l}{\textbf{Type-II: Biometrics and disease}}\\
  \cline{1-6}
  \multirow{3}{=}{Biometrics}
  & 3. MPI-LEMON-age \cite{babayan2019mind}
  & 203 & 64 & 1 s & Age group classification derived from the MPI-LEMON cohort (4 groups in the current mounted version). \\
  \cline{2-6}
  & 4. MPI-LEMON-gender  \cite{babayan2019mind}
  & 203 & 64 & 1 s & Gender classification derived from the MPI-LEMON cohort (2 classes). \\
  \cline{2-6}
  & 5. MPI-LEMON-extraversion  \cite{babayan2019mind}
  & 203 & 64 & 1 s & Extraversion classification (2 classes).\\
  \cline{1-6}
 
  \multirow{6}{=}{Epilepsy and abnormalities}
  & 6. HFO~\cite{ds003555:1.0.1}
  & 30 & 18 & 2 s & High-frequency oscillation related binary classification (2 classes).\\
  \cline{2-6}
  & 7. TUAB~\cite{10.3389/fnins.2016.00196}
  & 325 & 23 & 10 s & Clinical normal vs. abnormal EEG classification (2 classes).\\
  \cline{2-6}
  & 8. TUEV~\cite{harati2015tuev}
  & 370 & 32 & 5 s & EEG event classification (6 classes).\\
  \cline{2-6}
  & 9. TUEP~\cite{10.3389/fnins.2016.00196}
  & 200 & 32 & 10 s & Seizure-related binary classification (2 classes).\\
  \cline{2-6}
  & 10. TUSL~\cite{nedc_tuh_eeg_corpora}
  & 38 & 32 & 10 s & Sleep-state classification (3 classes).\\
  \cline{2-6}
  & 11. Siena EEG~\cite{detti2020siena}
  & 14 & 31 & 10 s & Seizure-related binary classification (2 classes).\\
  \cline{1-6}
  Neurodevelopmental disorders
  & 12. Adult ADHD \cite{bajestani2023dataset}
  & 121 & 64 & 2 s & Healthy vs.\ ADHD classification (2 classes).\\
  \cline{1-6}
  \multirow{3}{=}{Neurodegenerative disorders}
  & 13. AD65 ~\cite{miltiadous2023dataset}
  & 88 & 19 & 10 s & Neurodegenerative disease classification (3 classes).\\
  \cline{2-6}
  & 14. PD31 ~\cite{ds002778:1.0.5}
  & 31 & 64 & 1 s & Healthy vs.\ Parkinson's disease classification (2 classes).\\
  \cline{2-6}
  & 15. PD-Mortality ~\cite{ds007020:1.0.0}
  & 94 & 64 & 2 s & Mortality vs. survival classification (2 classes).\\
  \cline{1-6}
  \multirow{4}{=}{Mental disorders}
  & 16. MDD~\cite{Mumtaz2016}
  & 63 & 22 & 5 s & Healthy vs.\ major depressive disorder classification (2 classes).\\
  \cline{2-6}
  & 17. TDBRAIN~\cite{van2022two}
  & 285 & 26 & 2 s & Psychiatric phenotype classification (4 classes in the current mounted version).\\
  \cline{2-6}
  & 18. Depression resting~\cite{ds003478:1.1.0}
  & 122 & 67 & 1 s & Depression severity grouping via BDI.\\
  \cline{2-6}
  & 19. MODMA~\cite{cai2020modma}
  & 53 & 128 & 20 s & Depression / patient vs.\ control classification (2 classes).\\
  \midrule
  \addlinespace[3pt]
 
  \rowcolor{ourblue}\multicolumn{6}{l}{\textbf{Type-III: Consciousness and state}}\\
  \cline{1-6}
  Consciousness level detection
  & 20. Awakening~\cite{bajwa2025repeated}
  & 21 & 65 & 10 s & Awake vs.\ sedation state classification (2 classes).\\
  \cline{1-6}
  \multirow{5}{=}{Sleep staging}
  & 21. ISRUC-Sleep Subgroup I ~\cite{khalighi2016isruc}
  & 57 & 6 & 30 s & Sleep stage classification (5 classes).\\
  \cline{2-6}
  & 22. ISRUC-Sleep Subgroup II ~\cite{khalighi2016isruc}
  & 8 & 6 & 30 s & Sleep stage classification (5 classes).\\
  \cline{2-6}
  & 23. ISRUC-Sleep Subgroup III ~\cite{khalighi2016isruc}
  & 10 & 6 & 30 s & Sleep stage classification (5 classes).\\
  \cline{2-6}
  & 24. Sleep-EDF ~\cite{kemp2000analysis}
  & 153 & 2 & 30 s & Sleep stage classification (5 classes).\\
  \cline{2-6}
  & 25. HMC~\cite{alvarezestevez2022hmc}
  & 124 & 8 & 30 s & Sleep stage classification (5 classes).\\
  \cline{1-6}
  \multirow{3}{=}{Cognitive task identification}
  & 26. HBN-EEG ~\cite{shirazi2024hbn,alexander2017open}
  & 136 & 129 & 2 s & Multi-context task-type classification (13 classes in the current mounted version).\\
  \cline{2-6}
  & 27. PEARL-Neuro ~\cite{dzianok2024pearl}
  & 79 & 128 & 1 s & Context/task discrimination classification (3 classes in the current mounted version).\\
  \cline{2-6}
  & 28. RestCog ~\cite{ds002721:1.0.3}
  & 60 & 61 & 1 s & Task-type classification (5 classes).\\
  \midrule
  \addlinespace[3pt]
 
  \rowcolor{ourblue}\multicolumn{6}{l}{\textbf{Type-IV: Cognition and emotion}}\\
  \cline{1-6}
  Vigilance detection
  & 29. SEED-VIG ~\cite{zheng2017multimodal}
  & 21 & 17 & 8 s & Vigilance state classification (3 classes in the current mounted version).\\
  \cline{1-6}
  \multirow{12}{=}{Emotion recognition}
  & 30. DEAP-arousal ~\cite{koelstra2011deap}
  & 32 & 32 & 10 s & High/low arousal classification (2 classes).\\
  \cline{2-6}
  & 31. DEAP-valence ~\cite{koelstra2011deap}
  & 32 & 32 & 10 s & High/low valence classification (2 classes).\\
  \cline{2-6}
  & 32. FACED ~\cite{chen2023large}
  & 123 & 32 & 10 s & Fine-grained emotion classification (9 classes).\\
  \cline{2-6}
  & 33. SEED ~\cite{duan2013differential}
  & 15 & 62 & 10 s & Positive/negative/neutral video-elicited emotion classification (3 classes).\\
  \cline{2-6}
  & 34. SEED-IV ~\cite{zheng2018emotionmeter}
  & 15 & 62 & 4 s & Emotion classification (4 classes).\\
  \cline{2-6}
  & 35. SEED-V ~\cite{liu2021comparing}
  & 16 & 62 & 1 s & Audio-visual elicited emotion classification (5 classes).\\
  \cline{2-6}
  & 36. SEED-VII ~\cite{jiang2024seed}
  & 20 & 62 & 10 s & Audio-visual elicited emotion classification (7 classes).\\
  \cline{2-6}
  & 37. SEED-FRA ~\cite{imtiaz2015open}
  & 8 & 60 & 10 s & Emotion classification with French movie stimuli (3 classes).\\
  \cline{2-6}
  & 38. EEG-SvRec  ~\cite{zhang2024eeg}
  & 30 & 69 & 1 s & Affective / preference related binary classification (2 classes in the current mounted version).\\
  \cline{2-6}
  & 39. CIRE ~\cite{he2025cire}
  & 38 & 128 & 2 s & Speech-related affective classification (2 classes in the current mounted version).\\
  \cline{2-6}
  & 40. MusicEEG~\cite{ds002721:1.0.3}
  & 31 & 19 & 1 s & Music-evoked emotion classification (2 classes).\\
  \cline{2-6}
  & 41. EmoEEG-MC ~\cite{xu2025multi}
  & 53 & 64 & 5 s & Multi-context emotion classification.\\
  \cline{1-6}
  \multirow{2}{=}{Workload detection}
  & 42. EEGMAT ~\cite{zyma2019eegmat}
  & 36 & 21 & 5 s & Cognitive workload / task-state classification (2 classes).\\
  \cline{2-6}
  & 43. Workload ~\cite{hinss2023workload}
  & 12 & 61 & 2 s & Workload level classification (3 classes).\\
  \cline{1-6}
  \addlinespace[3pt]
 
  \rowcolor{ourblue} \multicolumn{6}{l}{\textbf{Type-V: Naturalistic stimulus decoding}}\\
  \cline{1-6}
  Imagined speech
  & 44. BCI-speech~\cite{bci2020_track3}
  & 45 & 64 & 3 s & Speech intention / keyword classification (5 classes).\\
  \cline{1-6}
  \multirow{2}{=}{Listening}
  & 45. Broderick (cocktail party)~\cite{broderick2018electrophysiological}
  & 33 & 69 & 2 s & Left--right auditory attention classification (2 classes).\\
  \cline{2-6}
  & 46. Broderick (reverse)~\cite{broderick2018electrophysiological}
  & 19 & 69 & 2 s & Natural vs.\ time-reversed speech classification (2 classes).\\
  \cline{1-6}
  Reading
  & 47. ChineseEEG2 ~\cite{chen2025eeg}
  & 4 & 128 & 2 s & Tone classification (4 classes).\\
  \cline{1-6}
  Visual
  & 48. Things-EEG2~\cite{gifford2022large}
  & 10 & 17 & 1 s & Animate vs.\ inanimate concept classification (2 classes).\\
  \cline{1-6}
  \addlinespace[3pt]
 
  \rowcolor{ourblue} \multicolumn{6}{l}{\textbf{Type-VI: Motor and interaction}}\\
  \cline{1-6}
  \multirow{4}{=}{Motor imagery}
  & 49. BCI Competition IV-1~\cite{blankertz2007non}
  & 7 & 59 & 4 s & Motor imagery classification (2 classes).\\
  \cline{2-6}
  & 50. BCI Competition IV-2A~\cite{brunner2008bci}
  & 9 & 22 & 4 s & Motor imagery classification (4 classes).\\
  \cline{2-6}
  & 51. PhysioNet-MI~\cite{schalk2004bci2000}
  & 109 & 64 & 4 s & Motor imagery classification (4 classes).\\
  \cline{2-6}
  & 52. SHU-MI~\cite{Ma2022}
  & 25 & 32 & 4 s & Motor imagery classification (2 classes).\\
  \cline{1-6}
  \multirow{4}{=}{SSVEP}
  & 53. BETA~\cite{liu2020beta}
  & 70 & 64 & 1 s & SSVEP target classification (40 classes).\\
  \cline{2-6}
  & 54. Benchmark-SSVEP ~\cite{wang2017benchmarkssvep}
  & 35 & 64 & 2 s & SSVEP target classification (40 classes).\\
  \cline{2-6}
  & 55. Dual-Freq-SSVEP~\cite{sun2024dualfreqssvep}
  & 14 & 64 & 1 s & Dual-frequency SSVEP target classification (40 classes).\\
  \cline{2-6}
  & 56. SSVEP-9-ch~\cite{openneuro_ds006940}
  & 20 & 9 & 3 s & SSVEP target classification (160 classes).\\
  \cline{1-6}
  ErrP feedback
  & 57. Monitoring ErrP~\cite{chavarriagamonitoring}
  & 6 & 64 & 1 s & Error-related potential classification (2 classes in the current mounted version).\\
  \cline{1-6}
  Closed-loop assistive control
  & 58. EEG-Controlled Exoskeleton~\cite{ds006940:1.0.0}
  & 7 & 60 & 1 s & Closed-loop control of walking and stopping (2 classes).\\
 
  \end{longtable}
  \endgroup

\section{The rationale of data windowing}
\label{sec:data_windowing}

Because OmniEEG-Bench integrates datasets with heterogeneous experimental designs, sample length was determined at the dataset level rather than fixed globally. When a dataset or clinical convention defines a standard epoch, we follow that convention. Sleep-staging datasets are segmented into 30-s epochs, consistent with standard sleep-scoring practice and the annotations used in ISRUC-Sleep and Sleep-EDF. EEGDenoiseNet is kept at its native 2-s segment length. For motor imagery datasets, we use the task-relevant imagery interval when available, such as the 4-s cue or imagery period in BCI Competition IV datasets. 

For datasets with longer continuous recordings but no single canonical trial length, we choose windows according to the dominant temporal scale of the downstream task. Resting-state and clinical-state datasets are segmented into windows long enough to capture stable spectral or state-level features while remaining approximately stationary. For example, MODMA uses longer windows (20s) following prior depression-classification practice with long resting-state segments~\cite{gulenc2024diagnosis}. For naturalistic language and speech datasets, such as ChineseEEG2 and Broderick, the window length (2s) is designed to retain local linguistic or prosodic context while avoiding excessive mixing across adjacent words, sentences, or stimulus events. For SSVEP and closed-loop BCI tasks, we use short windows to emphasize low-latency decoding. 

All segmentation choices are fixed before model evaluation and are applied identically to all models within each dataset.

\section{Benchmarked EEG foundation models}
\label{sec:eeg_fm}

We selected EEG-FMs to provide a representative and reproducible coverage of current foundation-model designs for EEG. Models were included if they provide accessible pretrained weights or reproducible implementations, are intended for general-purpose or cross-dataset EEG representation learning, and can be adapted to the unified OmniEEG-Bench evaluation interface. The selected models span several major methodological families: Transformer-based masked modeling, state-space or Mamba-based sequence modeling, contrastive biosignal representation learning, and multi-task or task-aligned pretraining. They also differ substantially in pretraining scale, data diversity, input modality, channel-processing strategy, and model size. This diversity allows OmniEEG-Bench to evaluate not only which model performs best, but also which design factors are associated with stronger transfer across heterogeneous EEG tasks. Supplementary Table 2 summarizes characteristics of EEG-FMs included in OmniEEG-Bench. Because papers often differ in how they report data volume and parameter counts (e.g., multiple model sizes or partially reported hours), we keep the table faithful to the original disclosures and mark unreported fields accordingly. The underlined parameter size is that of the "base" model and is used in our benchmark. To quantify the data diversity used in the pre-training of each model and exclude the training datasets from evaluation, we list the pre-training datasets for each model in Supplementary Table 3. To ensure reproducibility, we summarize the implementation details of each model in Supplementary Table 4, including the target sampling rate, input shape, channel input policy, and normalization strategy. Notably, models differ substantially in how they handle channel configurations: some accept arbitrary layouts via adaptive encoding (CBraMod) or coordinate-based mapping (BrainOmni), while others require fixed montages such as the 10--20 system (LaBraM, NeuroLM) or predefined bipolar derivations (FEMBA, Neuro-GPT, BIOT).

\section{Validity of the Downsampled Evaluation Protocol}
\label{sec:downsample_validity}

To justify the use of a downsampled evaluation protocol in linear probing, we provide 
two complementary analyses. We first determine an appropriate number of training samples 
per class by examining how performance and its variance evolve with increasing sample 
size. As shown in Supplementary Fig. 1, performance stabilizes 
beyond approximately 40 samples per class, after which variance across repeated 
subsampling runs also diminishes. This supports the choice of 40 samples per subject 
per class as a reliable operating point. We then verify that this downsampled protocol 
preserves the relative ranking of models compared to using the full dataset, as shown 
in Supplementary Fig. 2. Model rankings under both protocols 
remain largely consistent across task categories and evaluation settings, confirming 
that downsampling does not introduce systematic bias in comparative evaluation.

\begin{figure*}[!ht]
  \centering
  \includegraphics[width=0.85\textwidth]{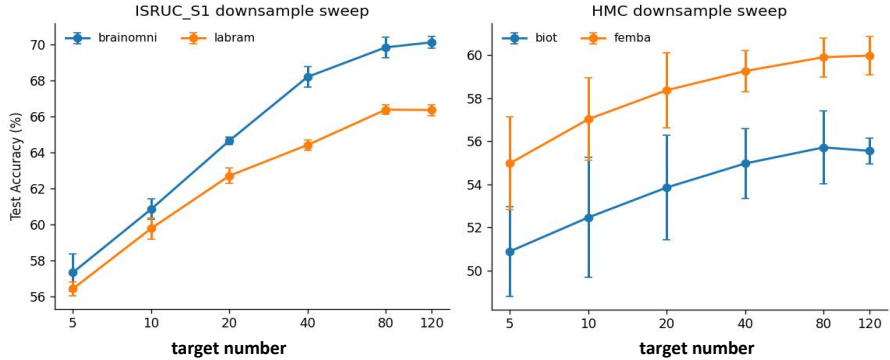}
  \caption{Supplementary Fig. 1. Model performance as a function of the number of 
  training samples per class under the downsampled linear probing setting, shown for 
  two representative datasets (ISRUC-S1 and HMC). Error bars denote standard deviation 
  across 15 random subsampling runs.}
  \label{fig:sample_size}
\end{figure*}

\begin{figure*}[!ht]
  \centering
  \includegraphics[width=1.03\textwidth]{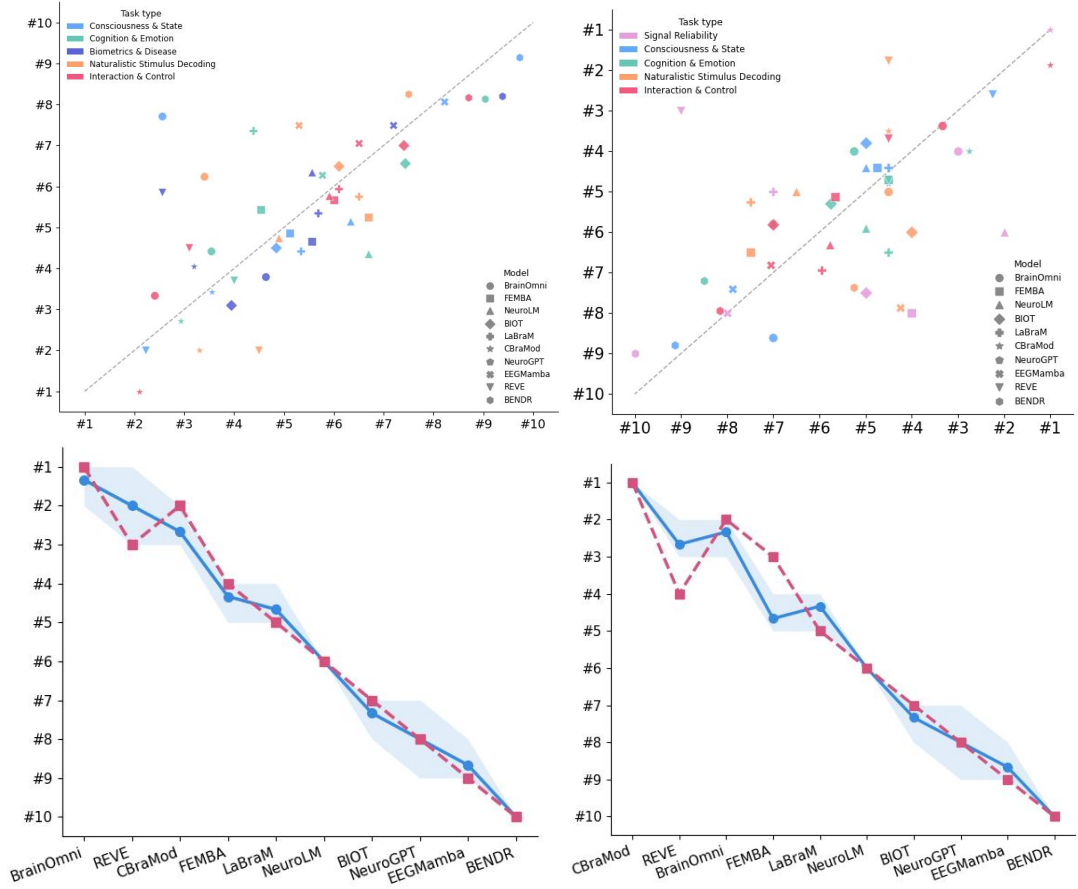}
  \caption{Supplementary Fig. 2. Rank consistency between full-sample and downsampled 
  evaluation settings. \textit{Top}: Scatter plots comparing model ranks under 
  cross-subject (left) and multi-subject (right) protocols, where each point encodes 
  a model (shape) and task category (color). \textit{Bottom}: Overall model ranking 
  curves under cross-subject (left) and multi-subject (right) protocols; the blue 
  solid line and shaded band denote the mean and min--max range across three seeds 
  under the downsampled setting, and the red dashed line denotes the full-sample rank.}
  \label{fig:full_vs_down}
\end{figure*}

\section{Full Fine-tuning Results}
\label{sec:full_finetune}

We report full fine-tuning results under the cross-subject setting in 
Supplementary Fig. 4. Note that these results are 
obtained under the same downsampled evaluation protocol (40 samples per 
subject per class) as described in Supplementary Section D. 
Compared to linear probing, model rankings shift substantially under full 
fine-tuning, with LaBraM, NeuroLM, and FEMBA achieving the top three 
overall ranks. Notably, EEGConformer, included as a task-specific baseline, 
ranks competitively among foundation models, highlighting the benefit of 
end-to-end optimization for task-specific adaptation.

\begin{figure*}[!ht] 
  \centering
  \includegraphics[width=1.03\textwidth]{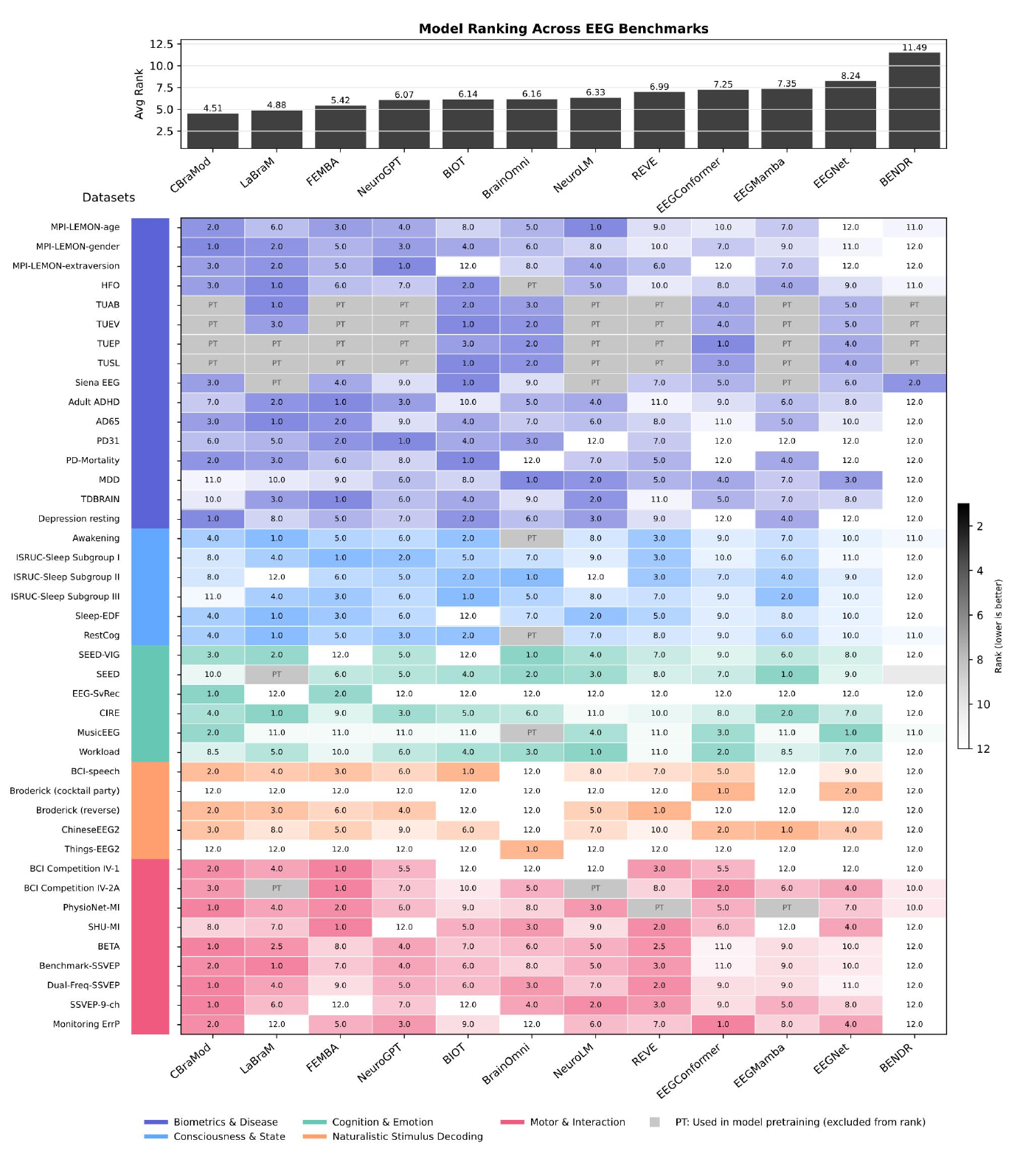}
  \caption{Supplementary Fig. 3. Full fine-tuning results under the 
  cross-subject setting. \textit{Top}: Average rank across all datasets 
  for each model. \textit{Bottom}: Per-dataset rank heatmap; lower rank 
  (darker color) indicates better performance. ``PT'' denotes that the 
  dataset was used in the model's pretraining and is excluded from ranking.}
  \label{fig:full_finetune}
\end{figure*}

\section{Multi-subject adaptation}

In the linear probing setting for multi-subject adaptation, CBraMod, REVE, and BrainOmni achieve the top three overall ranks with average ranks of 2.76, 3.25, and 4.36, respectively (Fig.~\ref{fig:within_subject}; see Supplementary Table 6 for detailed accuracy). The model ranking is broadly consistent with that observed in the cross-subject setting — where BrainOmni, CBraMod, and REVE also occupied the top three positions — with only minor reordering, suggesting that stronger pretrained representations tend to perform well under both cross-trial adaptation and cross-subject transfer. Several tasks remain difficult in the multi-subject setting. Eight models perform near chance on Chinese tone classification (ChineseEEG2-RA-Tone), and seven models perform near chance on workload classification (Workload), indicating that current EEG foundation models still have limited cross-trial discriminability for these tasks.

\begin{figure*}[h]
  \centering
  \includegraphics[width=1.03\linewidth]{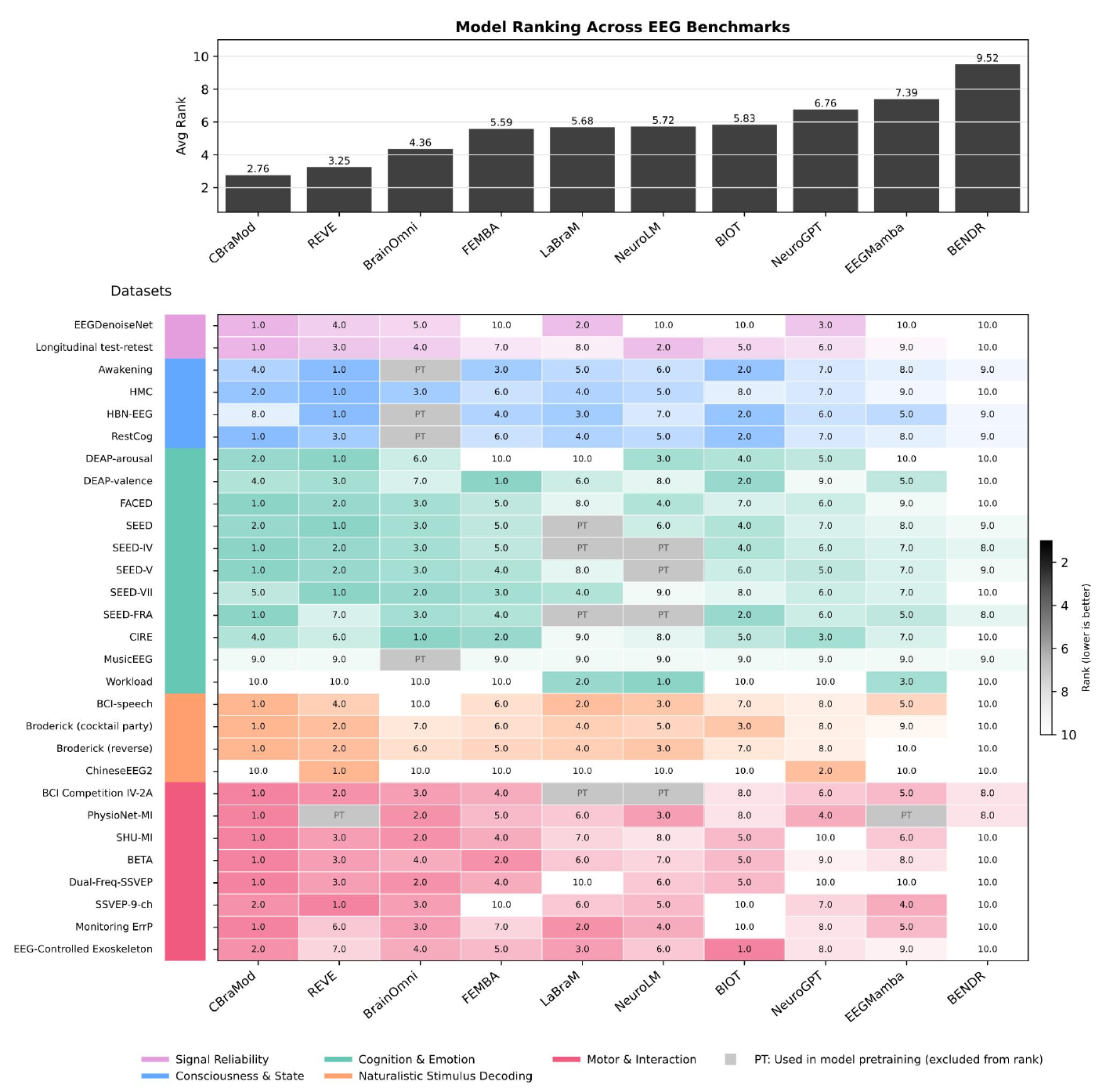}
  \caption{Supplementary Fig. 4. Multi-subject transfer ranks of ten EEG foundation models using OmniEEG-Bench.}
  \label{fig:within_subject}
\end{figure*}

\clearpage

\section{Comparison between Full Fine-tuning and Linear Probing}
\label{sec:full_finetune_Linear}

We compare model performance under full fine-tuning and linear probing in Supplementary Fig. 5. Task-specific baselines (EEGConformer and EEGNet, highlighted in blue) are included for reference.

Under full fine-tuning, the majority of foundation models outperform both task-specific baselines, suggesting that pretrained representations can be effectively adapted through end-to-end optimization. In contrast, under linear probing, the advantage of foundation models diminishes significantly. Many foundation models fail to match or surpass the task-specific baseline EEGConformer (average rank of 6.66), with only five top-performing models outperforming it. This clear performance drop indicates that while some advanced models show promise, many current EEG foundation models still struggle to provide sufficiently strong frozen representations that transfer across heterogeneous datasets without task-specific adaptation.

\begin{figure*}[!htbp]
  \centering
  \includegraphics[width=1.0\textwidth]{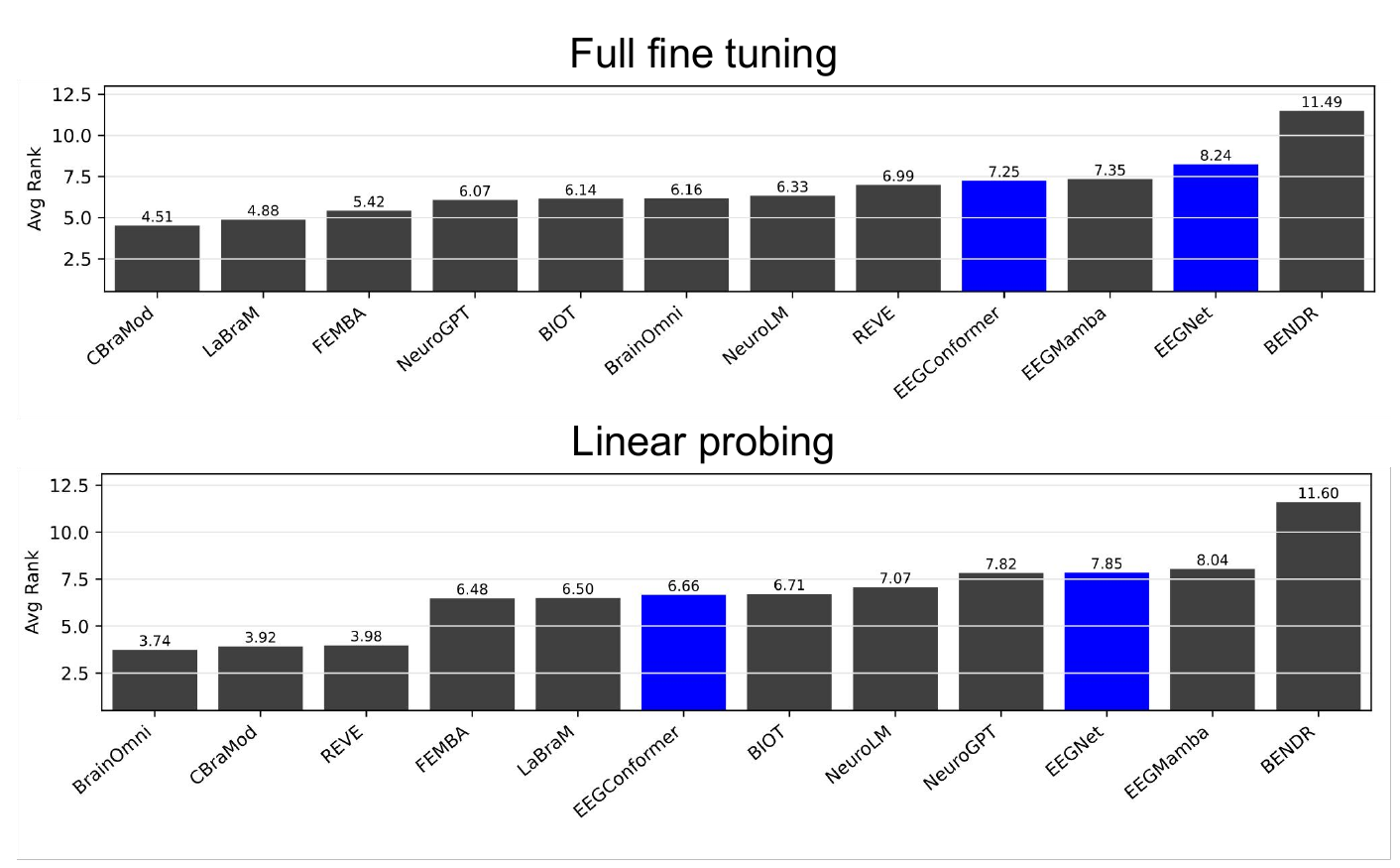}
j  \caption{Supplementary Fig. 5. Model ranks in full finetuning and linear probing.}
  \label{fig:model_rank_comparison}
\end{figure*}

\section{Average model ranks on each task taxonomy}
\label{sec:radar}

Supplementary Fig. 6 summarizes the average rank of each model across different task families under both evaluation protocols (where points closer to the outer edge indicate better ranks).

Under multi-subject adaptation (left), CBraMod and REVE exhibit the most consistently strong performance, forming the outermost boundaries of the radar chart. Specifically, CBraMod demonstrates exceptional strength in motor \& interaction and cognition \& emotion tasks, while REVE excels in signal reliability and naturalistic stimulus decoding.

Under cross-subject transfer (right), the performance landscape shifts, revealing distinct category-specific strengths. BrainOmni emerges as highly competitive, achieving top ranks in motor \& interaction and naturalistic stimulus decoding. Other models also display targeted advantages: BIOT notably spikes to the top rank in biometrics \& disease, REVE maintains dominance in consciousness \& state, and CBraMod remains a strong contender in cognition \& emotion.

Notably, no single model dominates uniformly across all task axes in either setting. The varying shapes of the radar webs emphasize that different model architectures capture complementary aspects of EEG represeantations, and their relative advantages depend heavily on the specific characteristics of the downstream task.

\begin{figure*}
  \centering
  \includegraphics[width=\textwidth]{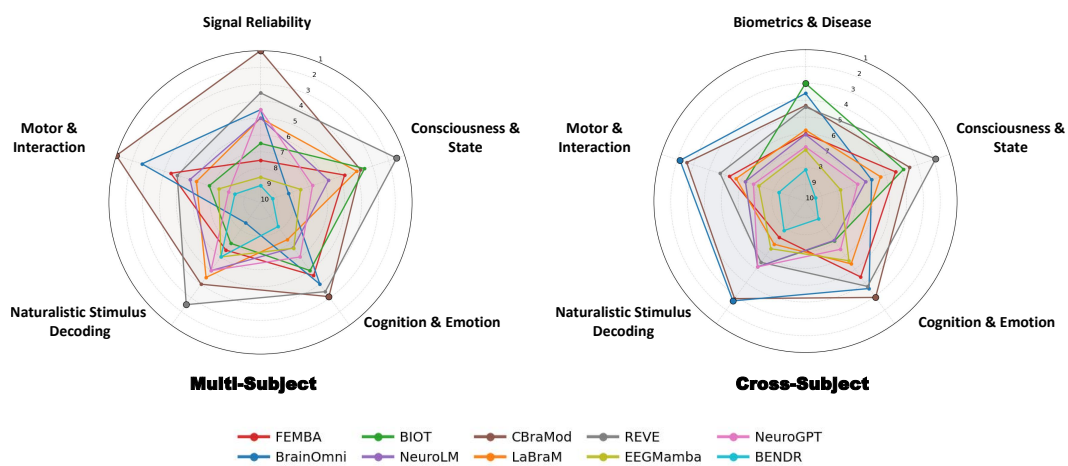}
  \caption{Supplementary Fig. 6. Average model ranks on each task taxonomy. Left: multi-subject adaptation; Right: cross-subject transfer.}
  \label{fig:radar}
\end{figure*}

\newcolumntype{L}[1]{>{\raggedright\arraybackslash}p{#1}}
\newcolumntype{C}[1]{>{\centering\arraybackslash}p{#1}}
\newcolumntype{Y}[1]{>{\raggedright\arraybackslash}p{#1}}

\clearpage
\begin{table*}[h]
\centering
\caption{\textit{Supplementary Table 2}. Summary of 10 EEG foundation models included in the benchmark.}
\label{tab:eeg_fm_summary_modified}
\small
\setlength{\tabcolsep}{3pt}
\renewcommand{\arraystretch}{1.12}
\begin{tabularx}{\textwidth}{@{}
L{2cm}
C{1.4cm}
C{1.8cm}
C{1.5cm}
L{1.2cm}
Y{2.0cm}
Y{1.6cm}
Y{2.1cm}
@{}}
\toprule
\textbf{Model} &
\makecell{\textbf{\#Training}\\\textbf{subj.}} &
\makecell{\textbf{Training data}\\\textbf{(h)}} &
\textbf{\#Params\footnotemark[1]} &
\textbf{Input} &
\textbf{Pre-training paradigm} &
\textbf{Backbone} &
\textbf{Channel processing} \\
\midrule
BrainOmni~\cite{xiao2025brainomni} & 6550 & 1,997 (EEG) + 656 (MEG) & 8.4M, \underline{37.7M} & EEG/MEG & Masked reconstruction (temporal) & Criss-Cross Transformer & Sensor Encoder \\
\midrule
FEMBA~\cite{tegon2025femba} & 14,987 & 27,062 & 7.8M, \underline{16.1M}, 77.8M, 389M & EEG & Masked reconstruction (temporal) & Mamba & fixed channel (10/20) \\
\midrule
NeuroLM~\cite{jiang2024neurolm} & 15,444 & 27,762 & \underline{255.1M}, 500M, 1.69B & EEG & Multi-task / autoregressive & Transformer & Unified channel vocabulary (10--20-based) \\
\midrule
CBraMod~\cite{wang2024cbramod} & 14,987 & 27,062 & \underline{5.0M} & EEG & Masked reconstruction (temporal)& Criss-Cross Transformer & Asymmetric Conditional Positional Encoding \\
\midrule
EEGMamba~\cite{wang2025eegmamba} & Unreported & 16,724 & \underline{3.3M} & EEG & Masked reconstruction (temporal)& Mamba & ST-Adaptive Module\\
\midrule
LaBraM~\cite{jiang2024large} &  Unreported & 2534.78 & \underline{5.8M} & EEG & Masked reconstruction (frequency) & Transformer & Unified channel vocabulary (10--20-based) \\
\midrule
Neuro-GPT~\cite{cui2024neuro} & 14,987 & 27,062 & \underline{78.4M} & EEG & Masked reconstruction (temporal) & GPT-2 & fixed channel (10/20) \\
\midrule
REVE~\cite{ouahidi2025reve} & 24,274 & 61,415 & \underline{69.2M} & EEG & Masked Autoencoder & Transformer & Unified channel embedding \\
\midrule
BIOT~\cite{yang2023biot} & Unreported & Unreported & \underline{3.2M} & EEG/ECG & Contrastive learning & Linear Transformer & fixed channel (18-channel bipolar montage) \\
\midrule
BENDR~\cite{kostas2021bendr} & 14,987 & 27,062 & \underline{4.0M} & EEG & Contrastive predictive coding & Transformer & fixed channel (10/20) \\
\bottomrule
\end{tabularx}
\footnotetext[1]{Parameter counts are reported in millions. Underlined numbers denote the total parameter count of the model instance used in our benchmark framework, as initialized by our \texttt{get\_model} implementation.}
\end{table*}

\begin{table*}[!htbp]
\centering
\caption{\textit{Supplementary Table 3}. Pre-training datasets of the foundation models.}
\label{tab:dataset_comparison}
\small
\begin{tabularx}{\textwidth}{@{}L{2.2cm}X@{}}
\toprule
\textbf{Model} & \textbf{Pre-training Dataset(s)} \\
\midrule
BrainOmni & \makecell[tl]{\textbf{EEG:} Go-Nogo, MusicEEG, HFO, SRM, RestCog, HBN EO/EC, Features-EEG,\\ PEARL-Neuro, HBN-EEG, Awakening\\\textbf{MEG:} MEG-MASC, MEG-Narrative-Dataset, OMEGA, CC700, AversiveMEG, MIND,\\ SMN4Lang, THINGS-MEG, ASWR-MEG, ImageLine, NeuroMorph, Kymata-SOTO. \\ \textbf{Total number: 22} } \\
\midrule
FEMBA & TUEG. \textbf{Total number: 1}\\
\midrule
NeuroLM & TUEG, SEED-IV, SEED-V, SEED-GER, SEED-FRA, BCI Competition IV-1, Emobrain, Grasp and Lift, Inria BCI, Motor Movement/Imagery, Raw EEG Data, Resting State, Siena Scalp EEG Database, SPIS Resting State, Target Versus Non-Target, Self-collected EEG corpus. \textbf{Total number: 16}   \\

\midrule
CBraMod & TUEG. \textbf{Total number: 1} \\
\midrule
EEGMamba & TUEG (clean subset) , PhysioNet 2018, Raw EEG Data, Siena Scalp EEG Database, B-SNIP1 (clean subset). \textbf{Total number: 5}\\
\midrule
LaBraM & \makecell[tl]{BCI Competition IV-1, Emobrain, Grasp and Lift EEG Challenge, Inria BCI Challenge, \\ EEG Motor Movement/Imagery Dataset, Raw EEG Data, Resting State EEG Data, SEED, \\ SEED-IV, SEED-GER, SEED-FRA, Siena Scalp EEG Database, SPIS Resting State Dataset, \\Target Versus Non-Target, TUAR, TUEP, TUSZ, TUSL, Self-collected EEG Data\\ \textbf{Total number: 19}}\\
\midrule
Neuro-GPT & TUEG. \textbf{Total number: 1} \\
\midrule
REVE & TUH, Physionet, OpenNeuro (), MOABB. \textbf{Total number: 92}\\ 
\midrule
BIOT & \makecell[tl]{\textbf{EEG:} SHHS, PREST\\\textbf{ECG:} Cardiology \\ \textbf{Total number: 3}}\\
\midrule
BENDR & TUEG. \textbf{Total number: 1}\\
\bottomrule
\end{tabularx}
\end{table*}

\begin{table*}[!htbp]
\centering
\caption{\textit{Supplementary Table 4}. Implementation details of the 10 EEG foundation models included in the benchmark.}
\label{tab:eeg_fm_impl}
\small
\setlength{\tabcolsep}{3pt}
\renewcommand{\arraystretch}{1.12}
\begin{tabularx}{\textwidth}{@{}
L{1.8cm}
C{1.4cm}
C{2.2cm}
L{3.5cm}
C{2.5cm}
C{1.8cm}
@{}}
\toprule
\textbf{Model} &
\makecell{\textbf{Target}\\\textbf{SR (Hz)}} &
\makecell{\textbf{Bandpass}\\\textbf{Filter (Hz)}} &
\textbf{Channel Policy} &
\textbf{Normalization} &
\makecell{\textbf{Sequence}\\\textbf{Length}} \\
\midrule
BrainOmni   & 256 & 0.1--96  & Maps input channels to standard electrode positions via 3D coordinates      & Z-score                  & Any \\
\midrule
FEMBA       & 200 & 0.1--75  & Interpolates to 22-channel TUEG bipolar montage via SSI                     & IQR normalization        & 1280 pts \\
\midrule
NeuroLM     & 200 & 0.1--75  & Retains channels matching the 10--20 system vocabulary                      & $\div$100 ($\mu$V to 0.1\,mV) & 200 pts/patch \\
\midrule
CBraMod     & 200 & 0.3--75  & Accepts arbitrary channel configurations via ACPE adaptive positional encoding & $\div$100 ($\mu$V to 0.1\,mV) & 200 pts/patch \\
\midrule
EEGMamba    & 200 & 0.1--50  & Reorders to 19-channel standard layout; missing channels zero-padded        & Identity                 & 6000 pts \\
\midrule
LaBraM      & 200 & 0.1--75  & Maps channels via 10--20 system vocabulary; requires channel ID as input    & $\div$100 ($\mu$V to 0.1\,mV) & 200 pts/patch \\
\midrule
Neuro-GPT   & 250 & 0.5--100 & Interpolates to 22-channel standard montage via SSI                         & Z-score                  & 500 pts/chunk \\
\midrule
REVE        & 256 & 0.5--99.5& Maps channels via unified channel embedding; accepts variable configurations & Z-score                  & Any \\
\midrule
BIOT        & 200 & --       & Constructs 18-channel bipolar montage (BIOT-18); missing channels zero-padded & P95 absolute scaling   & Any \\
\midrule
BENDR       & 256 & 0.1--100 & Maps to 19 standard channels plus one relative-amplitude auxiliary channel  & Identity (internal min-max) & Any \\
\bottomrule
\toprule
\end{tabularx}
\end{table*}

\section{Data preprocessing}
\label{sec:preproc}
As EEG signals are inherently noisy and complex, rigorous data preprocessing is critical to enhance signal quality and mitigate artifacts. To ensure consistency across heterogeneous datasets and facilitate efficient training of foundation models, we developed a unified HDF5-based benchmarking infrastructure. The standardized pipeline includes filtering, resampling, segmentation, and hierarchical storage.

\paragraph{Band-pass and notch filtering.}
EEG signals span a wide frequency range, but only specific bands are relevant to cognitive and emotional decoding tasks. We employ a standard 0.1--75 Hz band-pass filter using MNE-Python, preserving crucial task-relevant frequency components (from $\delta$ to $\gamma$ bands) while suppressing low-frequency drifts and high-frequency noise. In addition, a notch filter is dynamically configured (typically 60 Hz or 50 Hz) based on the data collection environment, effectively attenuating mains hum. The 0.1--75 band-pass filter is employed in the original implementation of a majority of EEG foundation models.

\paragraph{Resampling.}
Raw EEG recordings from different datasets are typically acquired at varying sampling rates (ranging from 128 Hz to 1000 Hz). We resample the EEG signals according to the requirement of each foundation model (250 Hz for Neuro-GPT, 256 Hz for BrainOmni and BENDR, and 200 Hz for others, Supplementary Table 4).

\paragraph{Windowing and segmentation.}
The data are segmented to generate fixed-length inputs for the model. We employ a sliding window strategy with a dataset-specific window ranging from 1s to 30s (non-overlapping). See Supplementary Table 1 for details.

\paragraph{Unified data storage.}
To resolve the disparity in raw data formats (e.g., .mat, .edf), all preprocessed data are organized into a standardized HDF5 schema. The root level stores subject-specific attributes (e.g., montage, sampling rate), trial groups organize recording sessions, and segment groups contain the actual preprocessed EEG matrices and synchronized labels, minimizing I/O latency while supporting flexible data splitting strategies during training. The data are split into train/val/test sets separately.

\section{Detailed results}

We report the means and standard deviations of balanced accuracy across three runs for linear probing of cross-subject transfer and multi-subject adaptation in Supplementary Tables 5 and 6, respectively. The cross-subject full-finetuning results are reported in Supplementary Table 7. In each table, \textbf{bold} entries indicate the best-performing model per dataset, \underline{underlined} entries indicate the second best, and \textit{italic} entries indicate performance below the chance level. ``--'' indicates that the model is incompatible with the dataset due to channel configuration constraints.


\begin{table*}[t]
\centering
\caption{\textit{Supplementary Table 5}. Cross-subject transfer results (balanced accuracy, mean\,$\pm$\,std over 3 runs).}
\label{tab:cross_results}
\scriptsize
\setlength{\tabcolsep}{3pt}
\renewcommand{\arraystretch}{1.05}
\resizebox{\textwidth}{!}{%
\begin{tabular}{lccccccccccc}
\toprule
Dataset & \#Classes & BrainOmni & CBraMod & REVE & FEMBA & BIOT & LaBraM & NeuroLM & NeuroGPT & EEGMamba & BENDR \\
\midrule
\multicolumn{12}{l}{\textbf{Type-I}} \\
Longitudinal test-retest & 2 & \textbf{0.755$\pm$0.006} & 0.743$\pm$0.020 & 0.725$\pm$0.025 & 0.683$\pm$0.038 & 0.677$\pm$0.005 & \underline{0.745$\pm$0.018} & 0.703$\pm$0.005 & 0.695$\pm$0.022 & 0.639$\pm$0.014 & \textit{0.500$\pm$0.000} \\
\midrule
\multicolumn{12}{l}{\textbf{Type-II}} \\
MPI-LEMON-age & 4 & 0.479$\pm$0.018 & 0.462$\pm$0.048 & \textbf{0.498$\pm$0.030} & \underline{0.480$\pm$0.025} & 0.477$\pm$0.033 & 0.439$\pm$0.048 & 0.403$\pm$0.052 & 0.426$\pm$0.046 & \textit{0.332$\pm$0.001} & 0.333$\pm$0.000 \\
MPI-LEMON-gender & 2 & 0.532$\pm$0.020 & \textbf{0.644$\pm$0.048} & \underline{0.599$\pm$0.030} & 0.569$\pm$0.058 & 0.587$\pm$0.060 & 0.571$\pm$0.073 & 0.553$\pm$0.052 & 0.545$\pm$0.055 & \textit{0.499$\pm$0.001} & \textit{0.500$\pm$0.000} \\
MPI-LEMON-extraversion & 2 & \textbf{0.536$\pm$0.013} & \textit{0.481$\pm$0.017} & 0.505$\pm$0.026 & \textit{0.490$\pm$0.012} & \textit{0.463$\pm$0.013} & 0.509$\pm$0.039 & \underline{0.518$\pm$0.026} & \textit{0.464$\pm$0.026} & 0.504$\pm$0.005 & \textit{0.500$\pm$0.000} \\
HFO & 2 & \textendash & 0.563$\pm$0.026 & \textbf{0.608$\pm$0.016} & 0.525$\pm$0.011 & \underline{0.586$\pm$0.029} & 0.541$\pm$0.015 & 0.537$\pm$0.009 & 0.544$\pm$0.010 & 0.580$\pm$0.022 & \textit{0.500$\pm$0.000} \\
TUAB & 2 & 0.727$\pm$0.034 & \textendash & \textendash & \textendash & \textbf{0.793$\pm$0.006} & \underline{0.732$\pm$0.009} & \textendash & \textendash & \textendash & \textendash \\
TUEV & 6 & \underline{0.467$\pm$0.034} & \textendash & \textendash & \textendash & \textbf{0.683$\pm$0.075} & 0.417$\pm$0.056 & \textendash & \textendash & \textendash & \textendash \\
TUEP & 2 & \textbf{0.596$\pm$0.053} & \textendash & \textendash & \textendash & \underline{0.588$\pm$0.031} & \textendash & \textendash & \textendash & \textendash & \textendash \\
TUSL & 3 & \underline{0.450$\pm$0.051} & \textendash & \textendash & \textendash & \textbf{0.645$\pm$0.055} & \textendash & \textendash & \textendash & \textendash & \textendash \\
Siena EEG & 2 & \underline{0.694$\pm$0.086} & \textbf{0.755$\pm$0.171} & 0.419$\pm$0.178 & 0.439$\pm$0.206 & 0.691$\pm$0.080 & \textendash & \textendash & 0.479$\pm$0.106 & \textendash & 0.333$\pm$0.236 \\
ADHD & 2 & 0.627$\pm$0.076 & \underline{0.666$\pm$0.077} & 0.626$\pm$0.059 & 0.602$\pm$0.018 & 0.535$\pm$0.035 & 0.580$\pm$0.084 & 0.662$\pm$0.057 & \textbf{0.673$\pm$0.110} & 0.552$\pm$0.082 & \textit{0.500$\pm$0.000} \\
AD65 & 3 & 0.462$\pm$0.111 & \textbf{0.582$\pm$0.130} & \underline{0.527$\pm$0.153} & 0.470$\pm$0.056 & 0.491$\pm$0.089 & 0.457$\pm$0.081 & 0.339$\pm$0.025 & 0.410$\pm$0.071 & 0.390$\pm$0.060 & 0.333$\pm$0.000 \\
PD31 & 2 & \textit{0.406$\pm$0.128} & \textit{0.369$\pm$0.167} & \textit{0.459$\pm$0.076} & \underline{0.511$\pm$0.140} & \textbf{0.544$\pm$0.129} & \textit{0.280$\pm$0.062} & \textit{0.414$\pm$0.126} & \textit{0.323$\pm$0.135} & \textit{0.336$\pm$0.136} & \textit{0.333$\pm$0.236} \\
PD-motaility & 2 & 0.501$\pm$0.003 & 0.530$\pm$0.016 & \underline{0.568$\pm$0.045} & \textit{0.500$\pm$0.010} & \textbf{0.623$\pm$0.049} & 0.516$\pm$0.017 & 0.560$\pm$0.075 & 0.530$\pm$0.046 & 0.501$\pm$0.003 & \textit{0.500$\pm$0.000} \\
MDD & 2 & \textbf{0.993$\pm$0.008} & 0.826$\pm$0.121 & \underline{0.948$\pm$0.063} & 0.742$\pm$0.058 & 0.861$\pm$0.075 & 0.743$\pm$0.133 & 0.883$\pm$0.090 & 0.640$\pm$0.055 & 0.701$\pm$0.045 & \textit{0.500$\pm$0.000} \\
TDBRAIN & 4 & 0.413$\pm$0.007 & \textbf{0.446$\pm$0.023} & \underline{0.437$\pm$0.019} & 0.396$\pm$0.018 & 0.393$\pm$0.030 & 0.391$\pm$0.027 & 0.413$\pm$0.006 & 0.362$\pm$0.008 & 0.323$\pm$0.021 & 0.278$\pm$0.039 \\
Depression-resting & 2 & 0.501$\pm$0.001 & 0.509$\pm$0.009 & \underline{0.511$\pm$0.000} & \textbf{0.522$\pm$0.017} & 0.505$\pm$0.006 & 0.500$\pm$0.000 & 0.503$\pm$0.002 & 0.503$\pm$0.003 & \textit{0.499$\pm$0.001} & \textit{0.500$\pm$0.000} \\
MODMA & 2 & \underline{0.657$\pm$0.152} & 0.500$\pm$0.000 & 0.572$\pm$0.076 & \textbf{0.773$\pm$0.072} & 0.500$\pm$0.000 & 0.500$\pm$0.000 & \textendash & 0.544$\pm$0.000 & 0.610$\pm$0.000 & 0.500$\pm$0.000 \\
\midrule
\multicolumn{12}{l}{\textbf{Type-III}} \\
Awakening & 2 & \textendash & 0.890$\pm$0.046 & \underline{0.962$\pm$0.026} & 0.961$\pm$0.026 & \textbf{0.973$\pm$0.019} & 0.934$\pm$0.041 & 0.811$\pm$0.023 & 0.781$\pm$0.089 & 0.833$\pm$0.024 & 0.500$\pm$0.000 \\
ISRUC-S1 & 5 & \underline{0.615$\pm$0.021} & 0.570$\pm$0.030 & \textbf{0.658$\pm$0.018} & 0.586$\pm$0.010 & 0.557$\pm$0.017 & 0.558$\pm$0.032 & 0.562$\pm$0.084 & 0.549$\pm$0.026 & 0.501$\pm$0.046 & 0.200$\pm$0.000 \\
ISRUC-S2 & 5 & \textbf{\textit{0.282$\pm$0.010}} & \textit{0.206$\pm$0.006} & \textit{0.249$\pm$0.023} & \textit{0.242$\pm$0.016} & \textit{0.250$\pm$0.014} & \textit{0.189$\pm$0.012} & \textit{0.207$\pm$0.008} & \underline{\textit{0.266$\pm$0.011}} & \textit{0.197$\pm$0.007} & \textit{0.200$\pm$0.000} \\
ISRUC-S3 & 5 & 0.345$\pm$0.014 & 0.319$\pm$0.021 & \textbf{0.374$\pm$0.025} & 0.321$\pm$0.010 & \underline{0.345$\pm$0.033} & 0.304$\pm$0.008 & 0.298$\pm$0.002 & 0.289$\pm$0.030 & 0.306$\pm$0.010 & \textit{0.200$\pm$0.000} \\
SleepEDF & 5 & 0.667$\pm$0.019 & \underline{0.690$\pm$0.016} & 0.681$\pm$0.009 & 0.625$\pm$0.014 & \textit{0.200$\pm$0.000} & \textbf{0.693$\pm$0.008} & 0.685$\pm$0.015 & 0.638$\pm$0.018 & 0.526$\pm$0.010 & \textit{0.200$\pm$0.000} \\
HMC & 5 & \underline{0.664$\pm$0.017} & 0.658$\pm$0.005 & \textbf{0.682$\pm$0.017} & 0.611$\pm$0.012 & 0.569$\pm$0.018 & 0.630$\pm$0.011 & 0.615$\pm$0.021 & 0.572$\pm$0.020 & 0.551$\pm$0.014 & \textit{0.200$\pm$0.000} \\
HBN EEG & 13 & \textendash & \underline{0.136$\pm$0.007} & \textbf{0.158$\pm$0.004} & 0.118$\pm$0.006 & 0.133$\pm$0.012 & 0.114$\pm$0.007 & \textit{0.090$\pm$0.004} & \textit{0.103$\pm$0.004} & \textit{0.100$\pm$0.003} & \textit{0.077$\pm$0.000} \\
PEARL-Neuro & 3 & \textendash & \textbf{0.535$\pm$0.006} & \underline{0.484$\pm$0.014} & 0.465$\pm$0.005 & 0.462$\pm$0.012 & 0.478$\pm$0.017 & 0.438$\pm$0.013 & 0.437$\pm$0.012 & 0.419$\pm$0.008 & 0.333$\pm$0.000 \\
RestCog & 5 & \textendash & \textbf{0.368$\pm$0.029} & 0.346$\pm$0.027 & 0.308$\pm$0.024 & \underline{0.350$\pm$0.025} & 0.318$\pm$0.023 & 0.309$\pm$0.024 & 0.276$\pm$0.017 & \textit{0.229$\pm$0.006} & \textit{0.200$\pm$0.000} \\
\midrule
\multicolumn{12}{l}{\textbf{Type-IV}} \\
SEED-VIG & 3 & 0.432$\pm$0.014 & \textbf{0.440$\pm$0.035} & 0.418$\pm$0.064 & 0.414$\pm$0.026 & 0.333$\pm$0.000 & \underline{0.437$\pm$0.025} & 0.360$\pm$0.028 & 0.432$\pm$0.028 & 0.350$\pm$0.059 & 0.333$\pm$0.000 \\
DEAP-arousal & 2 & 0.505$\pm$0.024 & \textbf{0.515$\pm$0.021} & 0.508$\pm$0.029 & 0.503$\pm$0.005 & 0.484$\pm$0.031 & 0.478$\pm$0.007 & 0.500$\pm$0.017 & \underline{0.510$\pm$0.011} & 0.487$\pm$0.025 & 0.500$\pm$0.000 \\
DEAP-valence & 2 & \underline{0.530$\pm$0.011} & 0.519$\pm$0.020 & 0.521$\pm$0.008 & 0.517$\pm$0.020 & 0.498$\pm$0.049 & 0.507$\pm$0.033 & 0.500$\pm$0.014 & 0.494$\pm$0.023 & \textbf{0.533$\pm$0.005} & 0.500$\pm$0.000 \\
SEED & 3 & \textit{0.492$\pm$0.013} & \underline{0.525$\pm$0.027} & \textbf{0.543$\pm$0.010} & \textit{0.434$\pm$0.008} & \textit{0.430$\pm$0.015} & \textendash & \textit{0.399$\pm$0.015} & \textit{0.466$\pm$0.035} & \textit{0.471$\pm$0.012} & \textit{0.333$\pm$0.000} \\
SEED-IV & 4 & \underline{0.316$\pm$0.017} & \textbf{0.334$\pm$0.005} & 0.297$\pm$0.013 & 0.303$\pm$0.009 & 0.259$\pm$0.029 & \textendash & \textendash & 0.292$\pm$0.006 & 0.288$\pm$0.008 & 0.250$\pm$0.000 \\
SEED-V & 5 & \textit{0.235$\pm$0.003} & \textit{0.234$\pm$0.008} & \underline{\textit{0.257$\pm$0.007}} & \textbf{\textit{0.290$\pm$0.006}} & \textit{0.257$\pm$0.010} & \textit{0.241$\pm$0.010} & \textendash & \textit{0.242$\pm$0.011} & \textit{0.204$\pm$0.006} & \textit{0.200$\pm$0.000} \\
SEED-VII & 7 & \textit{0.185$\pm$0.008} & \textit{0.176$\pm$0.009} & \textbf{\textit{0.191$\pm$0.006}} & \underline{\textit{0.189$\pm$0.015}} & \textit{0.159$\pm$0.006} & \textit{0.187$\pm$0.018} & \textit{0.161$\pm$0.005} & \textit{0.176$\pm$0.008} & \textit{0.181$\pm$0.013} & \textit{0.143$\pm$0.000} \\
SEED-FRA & 3 & \textbf{0.425$\pm$0.011} & \underline{0.417$\pm$0.013} & 0.353$\pm$0.006 & 0.398$\pm$0.023 & 0.385$\pm$0.013 & \textendash & \textendash & 0.366$\pm$0.014 & 0.409$\pm$0.003 & 0.333$\pm$0.000 \\
EEG-SVRec & 2 & \textbf{0.511$\pm$0.003} & 0.508$\pm$0.012 & \textit{0.499$\pm$0.002} & 0.506$\pm$0.002 & 0.503$\pm$0.003 & \underline{0.510$\pm$0.005} & 0.501$\pm$0.009 & \textit{0.500$\pm$0.004} & 0.504$\pm$0.005 & \textit{0.500$\pm$0.000} \\
FACED & 9 & \underline{\textit{0.204$\pm$0.002}} & \textbf{0.481$\pm$0.034} & \textit{0.203$\pm$0.016} & \textit{0.174$\pm$0.014} & \textit{0.156$\pm$0.006} & \textit{0.155$\pm$0.020} & \textit{0.170$\pm$0.022} & \textit{0.146$\pm$0.004} & \textit{0.151$\pm$0.011} & \textit{0.111$\pm$0.000} \\
MusicEEG & 2 & \textendash & \textbf{0.591$\pm$0.012} & \underline{0.528$\pm$0.059} & 0.486$\pm$0.023 & 0.475$\pm$0.014 & 0.524$\pm$0.032 & 0.519$\pm$0.031 & 0.480$\pm$0.039 & 0.528$\pm$0.045 & 0.500$\pm$0.000 \\
EmoEEG-MC & 10 & \textit{0.322$\pm$0.015} & 0.340$\pm$0.017 & \textit{0.305$\pm$0.035} & \textit{0.326$\pm$0.010} & \textit{0.332$\pm$0.017} & \underline{0.340$\pm$0.017} & \textit{0.327$\pm$0.009} & \textit{0.328$\pm$0.013} & \textbf{0.346$\pm$0.020} & 0.333$\pm$0.000 \\
CIRE & 2 & 0.531$\pm$0.017 & 0.526$\pm$0.018 & \textbf{0.538$\pm$0.017} & 0.532$\pm$0.029 & 0.513$\pm$0.012 & \underline{0.535$\pm$0.025} & 0.523$\pm$0.022 & 0.524$\pm$0.036 & 0.528$\pm$0.021 & \textit{0.500$\pm$0.000} \\
EEGMAT & 2 & \textit{0.485$\pm$0.022} & \textit{0.500$\pm$0.000} & \textit{0.499$\pm$0.013} & \underline{0.602$\pm$0.021} & 0.549$\pm$0.079 & \textit{0.495$\pm$0.007} & \textbf{0.616$\pm$0.082} & 0.522$\pm$0.035 & 0.503$\pm$0.004 & \textit{0.498$\pm$0.002} \\
Workload & 3 & 0.444$\pm$0.007 & 0.470$\pm$0.042 & 0.409$\pm$0.046 & 0.411$\pm$0.016 & 0.466$\pm$0.009 & \underline{0.478$\pm$0.028} & \textbf{0.496$\pm$0.019} & 0.358$\pm$0.023 & 0.406$\pm$0.022 & 0.333$\pm$0.000 \\
\midrule
\multicolumn{12}{l}{\textbf{Type-V}} \\
BCI-Speech & 5 & \underline{0.225$\pm$0.010} & \textbf{0.289$\pm$0.033} & 0.223$\pm$0.009 & 0.210$\pm$0.013 & 0.204$\pm$0.014 & 0.203$\pm$0.011 & 0.211$\pm$0.004 & 0.212$\pm$0.008 & 0.197$\pm$0.012 & 0.200$\pm$0.000 \\
Broderick-Cocktail-party & 2 & \textbf{0.516$\pm$0.004} & 0.384$\pm$0.052 & 0.399$\pm$0.068 & 0.358$\pm$0.039 & 0.346$\pm$0.085 & \underline{0.465$\pm$0.042} & 0.443$\pm$0.061 & 0.336$\pm$0.229 & 0.425$\pm$0.092 & 0.333$\pm$0.236 \\
Broderick-reverse & 2 & \textit{0.466$\pm$0.046} & \underline{0.580$\pm$0.049} & \textbf{0.710$\pm$0.138} & \textit{0.499$\pm$0.001} & \textit{0.481$\pm$0.026} & \textit{0.464$\pm$0.052} & \textit{0.492$\pm$0.058} & \textit{0.465$\pm$0.049} & 0.500$\pm$0.000 & \textit{0.500$\pm$0.000} \\
ChineseEEG2-RA-Tone & 4 & \textit{0.252$\pm$0.002} & \textbf{\textit{0.253$\pm$0.004}} & \textit{0.250$\pm$0.005} & \textit{0.250$\pm$0.001} & \underline{\textit{0.253$\pm$0.002}} & \textit{0.250$\pm$0.003} & \textit{0.252$\pm$0.001} & \textit{0.250$\pm$0.004} & \textit{0.252$\pm$0.001} & \textit{0.250$\pm$0.000} \\
ThingsEEG2 & 2 & \textbf{0.504$\pm$0.001} & \textit{0.500$\pm$0.000} & \textit{0.500$\pm$0.000} & \textit{0.500$\pm$0.000} & \textit{0.500$\pm$0.000} & \textit{0.500$\pm$0.000} & \textit{0.500$\pm$0.000} & \underline{0.502$\pm$0.002} & \textit{0.500$\pm$0.000} & \textit{0.500$\pm$0.000} \\
\midrule
\multicolumn{12}{l}{\textbf{Type-VI}} \\
BCIC-IV-2a & 4 & \underline{\textit{0.303$\pm$0.028}} & \textbf{\textit{0.329$\pm$0.006}} & \textit{0.298$\pm$0.009} & \textit{0.256$\pm$0.002} & \textit{0.250$\pm$0.000} & \textendash & \textendash & \textit{0.283$\pm$0.016} & \textit{0.289$\pm$0.009} & \textit{0.250$\pm$0.000} \\
BCIC-IV-1 & 2 & 0.510$\pm$0.011 & \textbf{0.552$\pm$0.031} & \underline{0.525$\pm$0.035} & 0.507$\pm$0.005 & 0.505$\pm$0.004 & 0.513$\pm$0.008 & 0.503$\pm$0.015 & \textit{0.463$\pm$0.009} & 0.508$\pm$0.010 & \textit{0.500$\pm$0.000} \\
Physionet-MI & 4 & \textit{0.307$\pm$0.011} & \textbf{\textit{0.499$\pm$0.007}} & \textendash & \textit{0.293$\pm$0.007} & \textit{0.256$\pm$0.008} & \textit{0.280$\pm$0.011} & \underline{\textit{0.312$\pm$0.016}} & \textit{0.305$\pm$0.016} & \textendash & \textit{0.250$\pm$0.000} \\
SHU-MI & 2 & \textbf{0.533$\pm$0.018} & 0.515$\pm$0.009 & \underline{0.523$\pm$0.003} & 0.511$\pm$0.002 & 0.519$\pm$0.014 & 0.507$\pm$0.011 & 0.504$\pm$0.012 & 0.504$\pm$0.002 & 0.490$\pm$0.012 & 0.500$\pm$0.000 \\
BETA-SSVEP & 40 & \textit{0.037$\pm$0.004} & \textbf{\textit{0.103$\pm$0.014}} & \underline{\textit{0.043$\pm$0.006}} & \textit{0.040$\pm$0.007} & \textit{0.031$\pm$0.004} & \textit{0.032$\pm$0.004} & \textit{0.033$\pm$0.004} & \textit{0.030$\pm$0.005} & \textit{0.028$\pm$0.008} & \textit{0.025$\pm$0.000} \\
Benchmark-SSVEP & 40 & \underline{0.078$\pm$0.021} & \textbf{0.425$\pm$0.103} & 0.056$\pm$0.008 & 0.051$\pm$0.009 & 0.065$\pm$0.019 & 0.033$\pm$0.003 & 0.032$\pm$0.002 & 0.026$\pm$0.005 & 0.033$\pm$0.006 & 0.025$\pm$0.000 \\
Dual-Freq-SSVEP & 40 & \underline{\textit{0.050$\pm$0.004}} & \textbf{\textit{0.079$\pm$0.006}} & \textit{0.035$\pm$0.005} & \textit{0.030$\pm$0.003} & \textit{0.023$\pm$0.006} & \textit{0.028$\pm$0.003} & \textit{0.027$\pm$0.005} & \textit{0.033$\pm$0.008} & \textit{0.032$\pm$0.003} & \textit{0.025$\pm$0.000} \\
SSVEP-9-chn & 160 & \textit{0.016$\pm$0.006} & \underline{\textit{0.029$\pm$0.002}} & \textbf{\textit{0.032$\pm$0.001}} & \textit{0.008$\pm$0.002} & \textit{0.006$\pm$0.000} & \textit{0.007$\pm$0.001} & \textit{0.013$\pm$0.004} & \textit{0.009$\pm$0.002} & \textit{0.011$\pm$0.002} & \textit{0.006$\pm$0.000} \\
Monitoring-Errp & 2 & \textbf{0.532$\pm$0.022} & 0.498$\pm$0.008 & 0.497$\pm$0.006 & 0.493$\pm$0.014 & 0.498$\pm$0.002 & \underline{0.518$\pm$0.014} & 0.487$\pm$0.004 & 0.482$\pm$0.014 & 0.510$\pm$0.014 & 0.500$\pm$0.000 \\
EEG-Controlled Exoskeleton & 2 & \textbf{0.523$\pm$0.001} & 0.513$\pm$0.021 & 0.509$\pm$0.016 & 0.502$\pm$0.009 & 0.488$\pm$0.057 & 0.492$\pm$0.016 & 0.518$\pm$0.028 & \underline{0.522$\pm$0.023} & 0.491$\pm$0.017 & 0.500$\pm$0.000 \\
\bottomrule
\end{tabular}%
}
\end{table*}

\begin{table*}[t]
\centering
\caption{\textit{Supplementary Table 6}. Multi-subject adaptation results (balanced accuracy, mean\,$\pm$\,std over 3 runs).}
\label{tab:multi_results}
\scriptsize
\setlength{\tabcolsep}{3pt}
\renewcommand{\arraystretch}{1.05}
\resizebox{\textwidth}{!}{%
\begin{tabular}{lccccccccccc}
\toprule
Dataset & \#Classes & CBraMod & REVE & BrainOmni & FEMBA & LaBraM & NeuroLM & BIOT & NeuroGPT & EEGMamba & BENDR \\
\midrule
\multicolumn{12}{l}{\textbf{Type-I}} \\
Longitudinal test-retest & 2 & \textbf{0.755$\pm$0.006} & 0.743$\pm$0.020 & 0.725$\pm$0.025 & 0.683$\pm$0.038 & 0.677$\pm$0.005 & \underline{0.745$\pm$0.018} & 0.703$\pm$0.005 & 0.695$\pm$0.022 & 0.639$\pm$0.014 & \textit{0.500$\pm$0.000} \\
\midrule
\multicolumn{12}{l}{\textbf{Type-III}} \\
Awakening & 2 & 0.888$\pm$0.049 & \textbf{0.976$\pm$0.008} & \textendash & 0.927$\pm$0.025 & 0.854$\pm$0.077 & 0.823$\pm$0.086 & \underline{0.964$\pm$0.005} & 0.776$\pm$0.067 & 0.707$\pm$0.066 & 0.500$\pm$0.000 \\
HMC & 5 & \underline{0.664$\pm$0.003} & \textbf{0.684$\pm$0.005} & 0.657$\pm$0.001 & 0.588$\pm$0.005 & 0.643$\pm$0.005 & 0.620$\pm$0.004 & 0.575$\pm$0.001 & 0.575$\pm$0.006 & 0.570$\pm$0.001 & \textit{0.200$\pm$0.000} \\
HBN EEG & 13 & \textit{0.087$\pm$0.004} & \textbf{0.138$\pm$0.006} & \textendash & \textit{0.108$\pm$0.009} & 0.117$\pm$0.006 & \textit{0.088$\pm$0.003} & \underline{0.120$\pm$0.005} & \textit{0.097$\pm$0.006} & \textit{0.108$\pm$0.006} & \textit{0.079$\pm$0.003} \\
RestCog & 5 & \textbf{0.413$\pm$0.001} & 0.362$\pm$0.004 & \textendash & 0.321$\pm$0.001 & 0.339$\pm$0.002 & 0.334$\pm$0.002 & \underline{0.367$\pm$0.002} & 0.293$\pm$0.001 & \textit{0.234$\pm$0.003} & \textit{0.200$\pm$0.000} \\
\midrule
\multicolumn{12}{l}{\textbf{Type-IV}} \\
DEAP-arousal & 2 & \underline{0.513$\pm$0.004} & \textbf{0.514$\pm$0.009} & 0.502$\pm$0.017 & 0.484$\pm$0.009 & 0.495$\pm$0.010 & 0.512$\pm$0.010 & 0.511$\pm$0.020 & 0.505$\pm$0.006 & 0.483$\pm$0.005 & 0.500$\pm$0.000 \\
DEAP-valence & 2 & 0.516$\pm$0.016 & 0.524$\pm$0.014 & 0.505$\pm$0.013 & \textbf{0.531$\pm$0.004} & 0.506$\pm$0.007 & 0.504$\pm$0.003 & \underline{0.524$\pm$0.004} & 0.504$\pm$0.021 & 0.509$\pm$0.009 & 0.500$\pm$0.000 \\
SEED & 3 & \underline{0.546$\pm$0.055} & \textbf{0.576$\pm$0.070} & 0.518$\pm$0.102 & \textit{0.457$\pm$0.129} & \textendash & \textit{0.438$\pm$0.041} & \textit{0.474$\pm$0.090} & \textit{0.433$\pm$0.039} & \textit{0.420$\pm$0.070} & \textit{0.333$\pm$0.000} \\
SEED-IV & 4 & \textbf{0.341$\pm$0.014} & \underline{0.340$\pm$0.023} & 0.305$\pm$0.013 & 0.300$\pm$0.020 & \textendash & \textendash & 0.303$\pm$0.006 & 0.287$\pm$0.002 & 0.278$\pm$0.005 & 0.250$\pm$0.000 \\
SEED-V & 5 & \textbf{\textit{0.263$\pm$0.032}} & \underline{\textit{0.251$\pm$0.029}} & \textit{0.243$\pm$0.018} & \textit{0.229$\pm$0.007} & \textit{0.211$\pm$0.033} & \textendash & \textit{0.216$\pm$0.005} & \textit{0.219$\pm$0.021} & \textit{0.214$\pm$0.013} & \textit{0.200$\pm$0.000} \\
SEED-VII & 7 & \textit{0.191$\pm$0.008} & \textbf{\textit{0.212$\pm$0.021}} & \underline{\textit{0.210$\pm$0.008}} & \textit{0.194$\pm$0.006} & \textit{0.191$\pm$0.009} & \textit{0.154$\pm$0.019} & \textit{0.164$\pm$0.006} & \textit{0.185$\pm$0.017} & \textit{0.175$\pm$0.011} & \textit{0.143$\pm$0.000} \\
SEED-FRA & 3 & \textbf{0.441$\pm$0.004} & 0.350$\pm$0.012 & 0.422$\pm$0.009 & 0.421$\pm$0.028 & \textendash & \textendash & \underline{0.440$\pm$0.007} & 0.372$\pm$0.016 & 0.376$\pm$0.001 & 0.333$\pm$0.000 \\
FACED & 9 & \textbf{0.484$\pm$0.023} & \underline{\textit{0.197$\pm$0.010}} & \textit{0.194$\pm$0.018} & \textit{0.170$\pm$0.004} & \textit{0.145$\pm$0.006} & \textit{0.171$\pm$0.016} & \textit{0.147$\pm$0.009} & \textit{0.151$\pm$0.004} & \textit{0.144$\pm$0.004} & \textit{0.111$\pm$0.000} \\
MusicEEG & 2 & 0.480$\pm$0.034 & 0.468$\pm$0.019 & \textendash & 0.458$\pm$0.024 & 0.465$\pm$0.015 & 0.470$\pm$0.020 & 0.462$\pm$0.022 & 0.453$\pm$0.039 & \underline{0.490$\pm$0.010} & \textbf{0.500$\pm$0.000} \\
CIRE & 2 & 0.545$\pm$0.014 & 0.535$\pm$0.028 & \textbf{0.554$\pm$0.021} & \underline{0.553$\pm$0.018} & 0.509$\pm$0.056 & 0.518$\pm$0.019 & 0.537$\pm$0.034 & 0.549$\pm$0.025 & 0.519$\pm$0.064 & \textit{0.500$\pm$0.000} \\
Workload & 3 & 0.315$\pm$0.032 & \textit{0.242$\pm$0.081} & 0.313$\pm$0.029 & 0.318$\pm$0.022 & \underline{0.335$\pm$0.002} & \textbf{0.356$\pm$0.024} & 0.297$\pm$0.053 & 0.259$\pm$0.073 & 0.334$\pm$0.002 & 0.333$\pm$0.000 \\
\midrule
\multicolumn{12}{l}{\textbf{Type-V}} \\
BCI-Speech & 5 & \textbf{0.321$\pm$0.011} & 0.225$\pm$0.013 & 0.198$\pm$0.022 & 0.214$\pm$0.014 & \underline{0.229$\pm$0.011} & 0.226$\pm$0.005 & 0.213$\pm$0.007 & 0.210$\pm$0.003 & 0.219$\pm$0.010 & 0.200$\pm$0.000 \\
Broderick-Cocktail-party & 2 & \textbf{0.804$\pm$0.006} & \underline{0.616$\pm$0.010} & 0.525$\pm$0.000 & 0.540$\pm$0.005 & 0.582$\pm$0.009 & 0.558$\pm$0.002 & 0.588$\pm$0.004 & 0.506$\pm$0.001 & 0.502$\pm$0.005 & 0.500$\pm$0.000 \\
Broderick-reverse & 2 & \textbf{0.838$\pm$0.017} & \underline{0.769$\pm$0.017} & 0.554$\pm$0.013 & 0.570$\pm$0.015 & 0.573$\pm$0.007 & 0.634$\pm$0.014 & 0.527$\pm$0.017 & 0.506$\pm$0.002 & \textit{0.500$\pm$0.001} & \textit{0.500$\pm$0.000} \\
ChineseEEG2-RA-Tone & 4 & \textit{0.248$\pm$0.007} & \textbf{\textit{0.253$\pm$0.002}} & \textit{0.247$\pm$0.010} & \textit{0.248$\pm$0.002} & \textit{0.245$\pm$0.004} & \textit{0.248$\pm$0.004} & \textit{0.248$\pm$0.001} & \underline{\textit{0.250$\pm$0.006}} & \textit{0.247$\pm$0.004} & \textit{0.250$\pm$0.000} \\
\midrule
\multicolumn{12}{l}{\textbf{Type-VI}} \\
BCIC-IV-2a & 4 & \textbf{\textit{0.401$\pm$0.020}} & \underline{\textit{0.335$\pm$0.005}} & \textit{0.308$\pm$0.005} & \textit{0.300$\pm$0.019} & \textendash & \textendash & \textit{0.250$\pm$0.000} & \textit{0.284$\pm$0.011} & \textit{0.291$\pm$0.005} & \textit{0.250$\pm$0.000} \\
Physionet-MI & 4 & \textbf{\textit{0.490$\pm$0.016}} & \textendash & \underline{\textit{0.291$\pm$0.010}} & \textit{0.264$\pm$0.015} & \textit{0.263$\pm$0.016} & \textit{0.285$\pm$0.026} & \textit{0.236$\pm$0.011} & \textit{0.279$\pm$0.016} & \textendash & \textit{0.250$\pm$0.000} \\
SHU-MI & 2 & \textbf{0.589$\pm$0.011} & 0.538$\pm$0.013 & \underline{0.557$\pm$0.027} & 0.535$\pm$0.019 & 0.506$\pm$0.015 & 0.501$\pm$0.013 & 0.528$\pm$0.008 & 0.497$\pm$0.012 & 0.507$\pm$0.003 & 0.500$\pm$0.000 \\
BETA-SSVEP & 40 & \textbf{\textit{0.097$\pm$0.003}} & \textit{0.037$\pm$0.006} & \textit{0.035$\pm$0.004} & \underline{\textit{0.038$\pm$0.006}} & \textit{0.029$\pm$0.003} & \textit{0.027$\pm$0.001} & \textit{0.033$\pm$0.003} & \textit{0.025$\pm$0.006} & \textit{0.026$\pm$0.001} & \textit{0.025$\pm$0.000} \\
Dual-Freq-SSVEP & 40 & \textbf{\textit{0.087$\pm$0.002}} & \textit{0.036$\pm$0.009} & \underline{\textit{0.052$\pm$0.007}} & \textit{0.032$\pm$0.005} & \textit{0.020$\pm$0.011} & \textit{0.025$\pm$0.005} & \textit{0.031$\pm$0.006} & \textit{0.023$\pm$0.003} & \textit{0.023$\pm$0.004} & \textit{0.025$\pm$0.000} \\
SSVEP-9-chn & 160 & \underline{\textit{0.023$\pm$0.006}} & \textbf{\textit{0.043$\pm$0.002}} & \textit{0.018$\pm$0.003} & \textit{0.006$\pm$0.001} & \textit{0.007$\pm$0.002} & \textit{0.011$\pm$0.003} & \textit{0.006$\pm$0.000} & \textit{0.007$\pm$0.003} & \textit{0.012$\pm$0.003} & \textit{0.006$\pm$0.000} \\
Monitoring-Errp & 2 & \textbf{0.540$\pm$0.014} & 0.503$\pm$0.016 & 0.509$\pm$0.008 & 0.502$\pm$0.013 & \underline{0.519$\pm$0.006} & 0.509$\pm$0.012 & 0.480$\pm$0.018 & 0.501$\pm$0.007 & 0.507$\pm$0.010 & 0.500$\pm$0.000 \\
EEG-Controlled Exoskeleton & 2 & \underline{0.544$\pm$0.004} & 0.516$\pm$0.004 & 0.519$\pm$0.009 & 0.519$\pm$0.008 & 0.528$\pm$0.003 & 0.518$\pm$0.003 & \textbf{0.546$\pm$0.003} & 0.513$\pm$0.018 & 0.505$\pm$0.005 & 0.500$\pm$0.000 \\
\bottomrule
\end{tabular}%
}
\end{table*}

\begin{table*}[t]
\centering
\caption{\textit{Supplementary Table 7}. Cross-subject full-finetuning results (balanced accuracy, mean\,$\pm$\,std over 3 runs).}
\scriptsize
\setlength{\tabcolsep}{3pt}
\renewcommand{\arraystretch}{1.05}
\resizebox{\textwidth}{!}{%
\begin{tabular}{lcccccccccccccc}
\toprule
Dataset & \#Classes & CBraMod & LaBraM & FEMBA & NeuroGPT & NeuroLM & BIOT & BrainOmni & REVE & EEGMamba & EEGConformer & EEGNet & BENDR \\
\midrule
\multicolumn{14}{l}{\textbf{Type-II}} \\
MPI-LEMON-age & 4 & \underline{0.581$\pm$0.051} & 0.518$\pm$0.031 & 0.540$\pm$0.010 & 0.537$\pm$0.019 & \textbf{0.582$\pm$0.030} & 0.485$\pm$0.008 & 0.527$\pm$0.007 & 0.457$\pm$0.021 & 0.502$\pm$0.004 & 0.343$\pm$0.028 & 0.333$\pm$0.000 & 0.333$\pm$0.000 \\
MPI-LEMON-gender & 2 & \textbf{0.692$\pm$0.022} & \underline{0.682$\pm$0.006} & 0.587$\pm$0.023 & 0.636$\pm$0.010 & 0.569$\pm$0.027 & 0.593$\pm$0.052 & 0.587$\pm$0.002 & 0.533$\pm$0.011 & 0.564$\pm$0.031 & 0.578$\pm$0.033 & 0.513$\pm$0.000 & \textit{0.500$\pm$0.000} \\
MPI-LEMON-extraversion & 2 & 0.552$\pm$0.010 & \underline{0.556$\pm$0.033} & 0.531$\pm$0.000 & \textbf{0.562$\pm$0.014} & 0.541$\pm$0.002 & \textit{0.463$\pm$0.000} & 0.502$\pm$0.000 & 0.508$\pm$0.017 & 0.506$\pm$0.015 & \textit{0.486$\pm$0.041} & \textit{0.471$\pm$0.033} & \textit{0.500$\pm$0.000} \\
HFO & 2 & 0.603$\pm$0.016 & \textbf{0.688$\pm$0.004} & 0.546$\pm$0.012 & 0.545$\pm$0.015 & 0.595$\pm$0.000 & \underline{0.647$\pm$0.030} & \textendash & 0.528$\pm$0.022 & 0.596$\pm$0.009 & 0.535$\pm$0.012 & 0.529$\pm$0.014 & \textit{0.500$\pm$0.000} \\
TUAB & 2 & \textendash & \textbf{0.767$\pm$0.009} & \textendash & \textendash & \textendash & \underline{0.761$\pm$0.016} & 0.703$\pm$0.021 & \textendash & \textendash & 0.687$\pm$0.037 & 0.666$\pm$0.015 & \textendash \\
TUEV & 6 & \textendash & 0.606$\pm$0.032 & \textendash & \textendash & \textendash & \textbf{0.707$\pm$0.014} & \underline{0.615$\pm$0.046} & \textendash & \textendash & 0.506$\pm$0.110 & 0.434$\pm$0.117 & \textendash \\
TUEP & 2 & \textendash & \textendash & \textendash & \textendash & \textendash & 0.623$\pm$0.022 & \underline{0.634$\pm$0.013} & \textendash & \textendash & \textbf{0.634$\pm$0.021} & 0.585$\pm$0.011 & \textendash \\
TUSL & 3 & \textendash & \textendash & \textendash & \textendash & \textendash & \textbf{0.742$\pm$0.066} & \underline{0.624$\pm$0.180} & \textendash & \textendash & 0.425$\pm$0.066 & \textit{0.287$\pm$0.005} & \textendash \\
Siena EEG & 2 & 0.742$\pm$0.041 & \textendash & 0.733$\pm$0.151 & 0.491$\pm$0.082 & \textendash & \textbf{0.795$\pm$0.033} & 0.445$\pm$0.129 & 0.573$\pm$0.022 & \textendash & 0.715$\pm$0.117 & 0.632$\pm$0.158 & \underline{0.750$\pm$0.250} \\
ADHD & 2 & 0.640$\pm$0.093 & \underline{0.718$\pm$0.099} & \textbf{0.728$\pm$0.017} & 0.714$\pm$0.127 & 0.713$\pm$0.093 & 0.593$\pm$0.004 & 0.702$\pm$0.055 & 0.565$\pm$0.035 & 0.645$\pm$0.108 & 0.593$\pm$0.060 & 0.627$\pm$0.022 & \textit{0.500$\pm$0.000} \\
AD65 & 3 & 0.612$\pm$0.048 & \textbf{0.631$\pm$0.136} & \underline{0.616$\pm$0.133} & 0.405$\pm$0.105 & 0.499$\pm$0.096 & 0.567$\pm$0.001 & 0.423$\pm$0.064 & 0.421$\pm$0.048 & 0.546$\pm$0.192 & 0.364$\pm$0.044 & 0.380$\pm$0.044 & 0.333$\pm$0.000 \\
PD31 & 2 & 0.563$\pm$0.000 & 0.585$\pm$0.000 & \underline{0.745$\pm$0.000} & \textbf{0.764$\pm$0.000} & \textit{0.098$\pm$0.000} & 0.602$\pm$0.000 & 0.651$\pm$0.000 & 0.537$\pm$0.000 & \textit{0.188$\pm$0.000} & \textit{0.377$\pm$0.190} & \textit{0.272$\pm$0.179} & \textit{0.000$\pm$0.000} \\
PD-motaility & 2 & \underline{0.664$\pm$0.106} & 0.642$\pm$0.033 & 0.545$\pm$0.040 & 0.506$\pm$0.008 & 0.518$\pm$0.018 & \textbf{0.683$\pm$0.047} & \textit{0.499$\pm$0.014} & 0.561$\pm$0.042 & 0.579$\pm$0.006 & \textit{0.499$\pm$0.003} & \textit{0.500$\pm$0.000} & \textit{0.500$\pm$0.000} \\
MDD & 2 & 0.736$\pm$0.259 & 0.773$\pm$0.173 & 0.775$\pm$0.145 & 0.885$\pm$0.048 & \underline{0.961$\pm$0.036} & 0.853$\pm$0.010 & \textbf{0.977$\pm$0.015} & 0.901$\pm$0.003 & 0.870$\pm$0.083 & 0.927$\pm$0.045 & 0.947$\pm$0.028 & \textit{0.500$\pm$0.000} \\
TDBRAIN & 4 & 0.370$\pm$0.002 & 0.421$\pm$0.009 & \textbf{0.448$\pm$0.044} & 0.385$\pm$0.011 & \underline{0.423$\pm$0.069} & 0.392$\pm$0.034 & 0.378$\pm$0.008 & 0.345$\pm$0.000 & 0.381$\pm$0.021 & 0.389$\pm$0.013 & 0.379$\pm$0.024 & \textit{0.250$\pm$0.000} \\
Depression-resting & 2 & \textbf{0.704$\pm$0.000} & 0.519$\pm$0.000 & 0.562$\pm$0.000 & 0.531$\pm$0.000 & 0.580$\pm$0.000 & \underline{0.613$\pm$0.000} & 0.537$\pm$0.000 & 0.519$\pm$0.000 & 0.572$\pm$0.000 & \textit{0.500$\pm$0.000} & \textit{0.500$\pm$0.000} & \textit{0.500$\pm$0.000} \\
\midrule
\multicolumn{14}{l}{\textbf{Type-III}} \\
Awakening & 2 & 0.979$\pm$0.010 & \textbf{0.992$\pm$0.008} & 0.975$\pm$0.006 & 0.975$\pm$0.009 & 0.893$\pm$0.051 & \underline{0.986$\pm$0.014} & \textendash & 0.982$\pm$0.014 & 0.955$\pm$0.033 & 0.805$\pm$0.001 & 0.668$\pm$0.008 & 0.500$\pm$0.000 \\
ISRUC-S1 & 5 & 0.586$\pm$0.026 & 0.653$\pm$0.051 & \textbf{0.691$\pm$0.001} & \underline{0.666$\pm$0.005} & 0.551$\pm$0.089 & 0.650$\pm$0.032 & 0.629$\pm$0.030 & 0.654$\pm$0.016 & 0.633$\pm$0.053 & 0.495$\pm$0.027 & 0.453$\pm$0.033 & 0.200$\pm$0.000 \\
ISRUC-S2 & 5 & \textit{0.218$\pm$0.002} & \textit{0.196$\pm$0.012} & \textit{0.235$\pm$0.015} & \textit{0.240$\pm$0.012} & \textit{0.199$\pm$0.004} & \underline{\textit{0.265$\pm$0.036}} & \textbf{\textit{0.276$\pm$0.000}} & \textit{0.242$\pm$0.013} & \textit{0.241$\pm$0.019} & \textit{0.227$\pm$0.024} & \textit{0.206$\pm$0.007} & \textit{0.200$\pm$0.000} \\
ISRUC-S3 & 5 & 0.233$\pm$0.011 & 0.327$\pm$0.010 & 0.333$\pm$0.064 & 0.304$\pm$0.020 & 0.298$\pm$0.048 & \textbf{0.355$\pm$0.039} & 0.326$\pm$0.001 & 0.303$\pm$0.021 & \underline{0.335$\pm$0.010} & 0.290$\pm$0.017 & 0.255$\pm$0.029 & \textit{0.200$\pm$0.000} \\
SleepEDF & 5 & 0.733$\pm$0.011 & \textbf{0.745$\pm$0.015} & 0.736$\pm$0.002 & 0.703$\pm$0.015 & \underline{0.740$\pm$0.006} & \textit{0.200$\pm$0.000} & 0.700$\pm$0.003 & 0.733$\pm$0.008 & 0.677$\pm$0.001 & 0.631$\pm$0.009 & 0.587$\pm$0.018 & \textit{0.200$\pm$0.000} \\
RestCog & 5 & 0.389$\pm$0.023 & \textbf{0.400$\pm$0.026} & 0.351$\pm$0.021 & 0.390$\pm$0.033 & 0.345$\pm$0.016 & \underline{0.396$\pm$0.030} & \textendash & 0.289$\pm$0.014 & 0.345$\pm$0.020 & 0.284$\pm$0.004 & 0.267$\pm$0.008 & \textit{0.200$\pm$0.000} \\
\midrule
\multicolumn{14}{l}{\textbf{Type-IV}} \\
SEED-VIG & 3 & 0.458$\pm$0.008 & \underline{0.479$\pm$0.031} & 0.321$\pm$0.065 & 0.406$\pm$0.013 & 0.439$\pm$0.012 & 0.333$\pm$0.000 & \textbf{0.505$\pm$0.007} & 0.392$\pm$0.003 & 0.396$\pm$0.025 & 0.340$\pm$0.008 & 0.342$\pm$0.000 & 0.333$\pm$0.000 \\
EEG-SVRec & 2 & \textbf{0.503$\pm$0.001} & \textit{0.497$\pm$0.003} & \underline{0.502$\pm$0.006} & \textit{0.498$\pm$0.000} & \textit{0.499$\pm$0.001} & \textit{0.499$\pm$0.005} & \textit{0.498$\pm$0.003} & \textit{0.497$\pm$0.003} & \textit{0.500$\pm$0.001} & \textit{0.498$\pm$0.004} & \textit{0.492$\pm$0.005} & \textit{0.500$\pm$0.000} \\
MusicEEG & 2 & \underline{0.556$\pm$0.027} & 0.454$\pm$0.008 & 0.492$\pm$0.067 & 0.485$\pm$0.031 & 0.532$\pm$0.074 & 0.378$\pm$0.068 & \textendash & 0.499$\pm$0.038 & 0.471$\pm$0.045 & 0.546$\pm$0.005 & \textbf{0.595$\pm$0.030} & 0.500$\pm$0.000 \\
CIRE & 2 & 0.524$\pm$0.006 & \textbf{0.551$\pm$0.007} & 0.515$\pm$0.010 & 0.528$\pm$0.041 & 0.506$\pm$0.017 & 0.523$\pm$0.013 & 0.523$\pm$0.009 & 0.511$\pm$0.001 & \underline{0.544$\pm$0.031} & 0.517$\pm$0.002 & 0.519$\pm$0.001 & \textit{0.500$\pm$0.000} \\
Workload & 3 & 0.400$\pm$0.011 & 0.437$\pm$0.020 & 0.395$\pm$0.013 & 0.404$\pm$0.010 & \textbf{0.492$\pm$0.021} & 0.447$\pm$0.037 & 0.473$\pm$0.018 & 0.360$\pm$0.003 & 0.400$\pm$0.015 & \underline{0.476$\pm$0.012} & 0.402$\pm$0.013 & 0.333$\pm$0.000 \\
\midrule
\multicolumn{14}{l}{\textbf{Type-V}} \\
BCI-Speech & 5 & \underline{0.262$\pm$0.010} & 0.225$\pm$0.003 & 0.249$\pm$0.003 & 0.210$\pm$0.022 & 0.207$\pm$0.005 & \textbf{0.267$\pm$0.025} & 0.199$\pm$0.007 & 0.208$\pm$0.012 & 0.200$\pm$0.008 & 0.222$\pm$0.022 & 0.206$\pm$0.009 & 0.200$\pm$0.000 \\
Broderick-Cocktail-party & 2 & 0.400$\pm$0.144 & 0.332$\pm$0.061 & 0.432$\pm$0.028 & 0.202$\pm$0.202 & 0.255$\pm$0.255 & 0.388$\pm$0.051 & 0.431$\pm$0.113 & 0.410$\pm$0.015 & 0.421$\pm$0.064 & \textbf{0.514$\pm$0.002} & \underline{0.503$\pm$0.004} & 0.250$\pm$0.250 \\
Broderick-reverse & 2 & \underline{0.688$\pm$0.116} & 0.656$\pm$0.035 & 0.566$\pm$0.067 & 0.570$\pm$0.051 & 0.570$\pm$0.029 & \textit{0.459$\pm$0.002} & \textit{0.443$\pm$0.035} & \textbf{0.918$\pm$0.049} & \textit{0.392$\pm$0.108} & \textit{0.496$\pm$0.003} & \textit{0.500$\pm$0.000} & \textit{0.500$\pm$0.000} \\
ChineseEEG2-RA-Tone & 4 & \textit{0.253$\pm$0.003} & \textit{0.251$\pm$0.003} & \textit{0.252$\pm$0.002} & \textit{0.251$\pm$0.001} & \textit{0.251$\pm$0.001} & \textit{0.252$\pm$0.002} & \textit{0.246$\pm$0.000} & \textit{0.250$\pm$0.001} & \textbf{\textit{0.255$\pm$0.007}} & \underline{\textit{0.253$\pm$0.005}} & \textit{0.252$\pm$0.003} & \textit{0.250$\pm$0.000} \\
ThingsEEG2 & 2 & \underline{\textit{0.500$\pm$0.000}} & \underline{\textit{0.500$\pm$0.000}} & \textit{0.498$\pm$0.002} & \underline{\textit{0.500$\pm$0.000}} & \underline{\textit{0.500$\pm$0.000}} & \textit{0.498$\pm$0.002} & \textbf{0.502$\pm$0.002} & \underline{\textit{0.500$\pm$0.000}} & \underline{\textit{0.500$\pm$0.000}} & \textit{0.498$\pm$0.002} & \textit{0.498$\pm$0.002} & \underline{\textit{0.500$\pm$0.000}} \\
\midrule
\multicolumn{14}{l}{\textbf{Type-VI}} \\
BCIC-IV-2a & 4 & \textit{0.372$\pm$0.005} & \textendash & \textbf{\textit{0.391$\pm$0.016}} & \textit{0.309$\pm$0.026} & \textendash & \textit{0.250$\pm$0.000} & \textit{0.338$\pm$0.011} & \textit{0.273$\pm$0.007} & \textit{0.324$\pm$0.010} & \underline{\textit{0.374$\pm$0.004}} & \textit{0.348$\pm$0.023} & \textit{0.250$\pm$0.000} \\
BCIC-IV-1 & 2 & \underline{0.517$\pm$0.022} & 0.507$\pm$0.028 & \textbf{0.527$\pm$0.033} & 0.502$\pm$0.013 & \textit{0.487$\pm$0.003} & \textit{0.458$\pm$0.043} & \textit{0.482$\pm$0.028} & 0.515$\pm$0.015 & \textit{0.490$\pm$0.000} & 0.502$\pm$0.007 & \textit{0.495$\pm$0.030} & \textit{0.500$\pm$0.000} \\
Physionet-MI & 4 & \textbf{0.549$\pm$0.000} & 0.501$\pm$0.022 & \underline{0.519$\pm$0.020} & \textit{0.480$\pm$0.007} & 0.508$\pm$0.004 & \textit{0.261$\pm$0.001} & \textit{0.387$\pm$0.017} & \textendash & \textendash & \textit{0.489$\pm$0.015} & \textit{0.437$\pm$0.016} & \textit{0.250$\pm$0.000} \\
SHU-MI & 2 & 0.520$\pm$0.018 & 0.525$\pm$0.009 & \textbf{0.544$\pm$0.002} & 0.500$\pm$0.002 & 0.502$\pm$0.000 & 0.527$\pm$0.013 & 0.536$\pm$0.021 & \underline{0.536$\pm$0.019} & 0.481$\pm$0.015 & 0.525$\pm$0.034 & 0.530$\pm$0.012 & 0.500$\pm$0.000 \\
BETA-SSVEP & 40 & \textbf{\textit{0.150$\pm$0.009}} & \underline{\textit{0.100$\pm$0.010}} & \textit{0.041$\pm$0.000} & \textit{0.095$\pm$0.005} & \textit{0.082$\pm$0.003} & \textit{0.044$\pm$0.009} & \textit{0.081$\pm$0.001} & \underline{\textit{0.100$\pm$0.022}} & \textit{0.037$\pm$0.005} & \textit{0.027$\pm$0.000} & \textit{0.028$\pm$0.001} & \textit{0.025$\pm$0.000} \\
Benchmark-SSVEP & 40 & \underline{0.678$\pm$0.090} & \textbf{0.701$\pm$0.096} & 0.291$\pm$0.078 & 0.617$\pm$0.090 & 0.340$\pm$0.065 & 0.310$\pm$0.048 & 0.268$\pm$0.011 & 0.674$\pm$0.140 & 0.116$\pm$0.002 & 0.036$\pm$0.004 & 0.062$\pm$0.006 & 0.025$\pm$0.000 \\
Dual-Freq-SSVEP & 40 & \textbf{\textit{0.173$\pm$0.027}} & \textit{0.057$\pm$0.012} & \textit{0.033$\pm$0.005} & \textit{0.052$\pm$0.002} & \textit{0.037$\pm$0.007} & \textit{0.043$\pm$0.010} & \textit{0.067$\pm$0.013} & \underline{\textit{0.089$\pm$0.009}} & \textit{0.033$\pm$0.002} & \textit{0.033$\pm$0.005} & \textit{0.027$\pm$0.004} & \textit{0.025$\pm$0.000} \\
SSVEP-9-chn & 160 & \textbf{0.227$\pm$0.001} & \textit{0.016$\pm$0.004} & \textit{0.005$\pm$0.002} & \textit{0.015$\pm$0.001} & \underline{\textit{0.140$\pm$0.026}} & \textit{0.006$\pm$0.000} & \textit{0.043$\pm$0.010} & \textit{0.110$\pm$0.009} & \textit{0.020$\pm$0.009} & \textit{0.008$\pm$0.001} & \textit{0.010$\pm$0.000} & \textit{0.006$\pm$0.000} \\
Monitoring-Errp & 2 & \underline{0.527$\pm$0.006} & 0.491$\pm$0.015 & 0.510$\pm$0.006 & 0.515$\pm$0.003 & 0.509$\pm$0.006 & 0.501$\pm$0.007 & 0.493$\pm$0.017 & 0.507$\pm$0.008 & 0.502$\pm$0.006 & \textbf{0.542$\pm$0.038} & 0.513$\pm$0.019 & 0.500$\pm$0.000 \\
\bottomrule
\end{tabular}%
}
\label{tab:supp6_cross_finetune}
\end{table*}

\section{Regression Tasks}
We evaluate on SEED-VIG, a vigilance estimation dataset containing 20{,}355 EEG segments 
from 21 subjects, where the label is a per-segment continuous value (ratio of eye closure, 
range $[0.02, 1.00]$). As shown in Table~\ref{tab:regression_seedvig}, most models yield 
near-zero or negative $R^2$ with high variance across seeds, indicating that frozen EEG 
representations generalize poorly to cross-subject prediction with continuous labels.

\begin{table}[h]
\centering
\caption{\textit{Supplementary Table 8.} Linear probing on SEED-VIG vigilance regression (cross-subject, mean\,$\pm$\,std over 3 seeds).}
\label{tab:regression_seedvig}
\setlength{\tabcolsep}{6pt}
\small
\begin{tabular}{l ccc}
\toprule
\textbf{Model} & $R^2\uparrow$ & $r\uparrow$ & RMSE$\downarrow$ \\
\midrule
BENDR     & $-0.009_{\pm.008}$ & $0.042_{\pm.038}$ & $0.257_{\pm.039}$ \\
BIOT      & $-0.064_{\pm.080}$ & $-0.082_{\pm.008}$ & $0.270_{\pm.032}$ \\
BrainOmni & $-0.073_{\pm.277}$ & $0.428_{\pm.239}$ & $0.259_{\pm.047}$ \\
CBraMod   & $-0.023_{\pm.433}$ & $0.548_{\pm.207}$ & $0.242_{\pm.028}$ \\
EEGMamba  & $-0.002_{\pm.272}$ & $0.463_{\pm.139}$ & $0.247_{\pm.025}$ \\
FEMBA     & $+0.054_{\pm.190}$ & $0.441_{\pm.167}$ & $0.242_{\pm.032}$ \\
LaBraM    & $-0.033_{\pm.340}$ & $0.458_{\pm.207}$ & $0.252_{\pm.032}$ \\
NeuroGPT  & $+0.076_{\pm.100}$ & $0.443_{\pm.204}$ & $0.238_{\pm.043}$ \\
NeuroLM   & $-0.019_{\pm.335}$ & $0.550_{\pm.113}$ & $0.249_{\pm.021}$ \\
REVE      & $-0.296_{\pm.905}$ & $0.559_{\pm.167}$ & $0.270_{\pm.059}$ \\
\bottomrule
\end{tabular}
\end{table}

\end{document}